\newif\ifqabg
\qabgtrue        

\ifqabg
  \PassOptionsToPackage{table,dvipsnames}{xcolor}
\else
  \PassOptionsToPackage{table}{xcolor}
\fi

\ifqabg
  \documentclass[10pt, a4paper, twocolumn, logo]{googledeepmind}
\else
  \documentclass[11pt]{article}
  \usepackage[review]{acl}
\fi

\usepackage{times}
\usepackage{microtype}
\usepackage{inconsolata}

\ifqabg
  \DeclareSymbolFontAlphabet{\mathbb}{AMSb}
\else
  \usepackage{latexsym}
  \usepackage[T1]{fontenc}
  \usepackage[utf8]{inputenc}
  \usepackage{pifont}
  \usepackage{graphicx}
  \usepackage{colortbl}
  \usepackage{booktabs}
  \usepackage{enumitem}
  \usepackage{url}
  \usepackage{textcomp}
\fi

\usepackage{amsmath,amsfonts,bm}
\usepackage{natbib}
\usepackage{fancyhdr}

\def\eqref#1{equation~\ref{#1}}

\def\1{\bm{1}}

\DeclareMathAlphabet{\mathsfit}{\encodingdefault}{\sfdefault}{m}{sl}
\SetMathAlphabet{\mathsfit}{bold}{\encodingdefault}{\sfdefault}{bx}{n}

\usepackage{xurl}
\usepackage{subcaption}
\usepackage{amsmath,amsfonts}
\usepackage[most]{tcolorbox}
\tcbuselibrary{breakable}
\usepackage{siunitx}
\usepackage{color}
\usepackage{wrapfig}
\usepackage{array}
\usepackage{stfloats}
\usepackage{float}
\usepackage{verbatim}
\usepackage{listings}
\usepackage[ruled, linesnumbered, lined]{algorithm2e}
\usepackage{multirow}
\usepackage{etoolbox}
\usepackage[normalem]{ulem}

\definecolor{BestColor}{HTML}{C8E6C9}
\definecolor{SecondBestColor}{HTML}{FFF9C4}
\definecolor{impcolor}{HTML}{2E8B57}
\definecolor{mygray}{gray}{0.9}

\definecolor{codebg}{HTML}{FAFBFC}        
\definecolor{codekw}{HTML}{005CC5}        
\definecolor{codestr}{HTML}{C7254E}       
\definecolor{codecomment}{HTML}{6A737D}   
\definecolor{codenum}{HTML}{D0D7DE}       
\definecolor{codeframe}{HTML}{D0D7DE}     
\definecolor{skillbg}{HTML}{F6F8FA}       
\definecolor{skillhdr}{HTML}{1F3D7A}      
\definecolor{skillem}{HTML}{6A0DAD}       

\lstdefinestyle{pycode}{
  language=Python,
  basicstyle=\ttfamily\scriptsize,
  keywordstyle=\color{codekw}\bfseries,
  stringstyle=\color{codestr},
  commentstyle=\color{codecomment}\itshape,
  numbers=left,
  numberstyle=\ttfamily\tiny\color{codenum},
  numbersep=6pt,
  xleftmargin=2.2em,
  backgroundcolor=\color{codebg},
  breaklines=true,
  breakatwhitespace=true,
  breakautoindent=false,
  breakindent=0pt,
  prebreak={},
  postbreak={},
  columns=fullflexible,
  keepspaces=true,
  showstringspaces=false,
  tabsize=2,
  upquote=true,
  literate={->}{$\rightarrow$}{2}
}

\definecolor{refgoodbg}{HTML}{C7E9C0}   
\definecolor{refbadbg}{HTML}{FBD3D3}    

\newcommand{\hlgood}[1]{{\setlength{\fboxsep}{1.2pt}\colorbox{refgoodbg}{\strut #1}}}
\newcommand{\hlbad}[1]{{\setlength{\fboxsep}{1.2pt}\colorbox{refbadbg}{\strut #1}}}

\lstdefinestyle{mdskill}{
  basicstyle=\ttfamily\scriptsize,
  backgroundcolor=\color{skillbg},
  morecomment=[l]{\#\#},
  commentstyle=\color{skillhdr}\bfseries,
  morekeywords={When-to-Use,Strategy,Common,Pitfalls,Visual,Anchor,References,Skill,Trigger},
  keywordstyle=\color{skillem}\bfseries,
  breaklines=true,
  breakatwhitespace=true,
  breakautoindent=false,
  breakindent=0pt,
  prebreak={},
  postbreak={},
  columns=fullflexible,
  keepspaces=true,
  showstringspaces=false,
  xleftmargin=4pt,
  literate=%
    {crop_good.png}{{\hlgood{\texttt{crop\_good.png}}}}{13}%
    {crop_bad.png}{{\hlbad{\texttt{crop\_bad.png}}}}{12}%
}

\lstdefinestyle{appendixsnippet}{
  basicstyle=\ttfamily\scriptsize,
  breaklines=true,
  breakatwhitespace=true,
  breakautoindent=false,
  breakindent=0pt,
  prebreak={},
  postbreak={},
  columns=fullflexible,
  keepspaces=true,
  showstringspaces=false
}

\tcbset{
  casebox/.style={
    enhanced, breakable,
    colback=white,
    colframe=codeframe,
    colbacktitle=gray!55,
    coltitle=white,
    fonttitle=\small\bfseries,
    boxrule=0.5pt,
    arc=3pt,
    titlerule=0pt,
    left=4pt, right=4pt, top=4pt, bottom=4pt,
  }
}

\newcommand{\sysname}{\textsc{Dynamo}}

\newtcolorbox{examplebox}[2][]{
  breakable, enhanced,
  colback=white, colframe=cyan, coltitle=white,
  fonttitle=\bfseries, title=#2,
  overlay middle={\draw[cyan, line width=1pt](frame.south west)--(frame.south east);},
  overlay last={\draw[cyan, line width=1pt](frame.south west)--(frame.south east);},
  #1
}

\newcommand{\improvementstyle}[1]{$^{\textcolor{impcolor}{\tiny #1}}$}
\newcommand{\scoreimp}[2]{%
  \textbf{#1}%
  \ifstrequal{#2}{+0.0}{}{%
    \ifstrequal{#2}{0.0}{}{%
      \makebox[0pt][l]{\improvementstyle{#2}}%
    }%
  }%
}

\title{\sysname{}: Dynamic Skill-Tool Evolution for Vision-Language Agents}

\ifqabg
  \author[1,2]{Yutao Sun\textsuperscript{*}}
  \author[1,3,4]{Yanting Miao\textsuperscript{*}}
  \author[1,5]{Hao-Xuan Ma\textsuperscript{*}}
  \author[1]{Mengyu Zhou\textsuperscript{$\dag$}}
  \author[2]{Mingshuai Chen}
  \author[6]{Tiancheng Zhao}
  \author[1]{Dexin Wang}
  \author[1]{Lei Lv}
  \author[1]{Li Xu}
  \author[1]{Xiaoxi Jiang}
  \author[1]{Guanjun Jiang}
  \affil[1]{Qwen Large Model Application Team, Alibaba}
  \affil[2]{Zhejiang University}
  \affil[3]{University of Waterloo}
  \affil[4]{Vector Institute}
  \affil[5]{Nanjing University}
  \affil[6]{Binjiang Institute of Zhejiang University}
  \correspondingauthor{Mengyu Zhou (\texttt{zhoumengyu.zmy@alibaba-inc.com})}
\else
  \author{Anonymous ACL submission}
\fi

\newcommand{\paperabstract}{%
Improving vision-language models (VLMs) on visual reasoning typically requires retraining
or hand-designed prompts and tools. We present \sysname{}, a training-free framework
that adapts a frozen VLM without any weight updates. On a small labeled training
subset, the agent inspects its own correct and incorrect attempts and evolves two
complementary capabilities: reusable reasoning skills for cognitive bottlenecks, and
executable visual tools for perceptual ones. Each generated tool is paired with a skill
that specifies when to invoke it, and both capability types accumulate in a persistent
library.
Across four visual reasoning benchmarks and five VLM backbones, \sysname{} improves direct
inference on all 20 model--benchmark settings (avg.\ $+5.6$ acc). When the tool set is given in advance, the framework learns when to call each tool, and
per-step tool choice improves on every tested backbone. Against task-specific RL (VTool-R1, DeepEyes), \sysname{} closes 65--99\% of the RL gap at
a fraction of the compute, and combines additively with RL when available.%
}

\ifqabg
  \begin{abstract}\paperabstract\end{abstract}
\fi

\begin{document}
\maketitle

\ifqabg\else
  \begin{abstract}\paperabstract\end{abstract}
\fi

\section{Introduction}

\begin{figure}[t]
  \centering
  \includegraphics[width=\columnwidth]{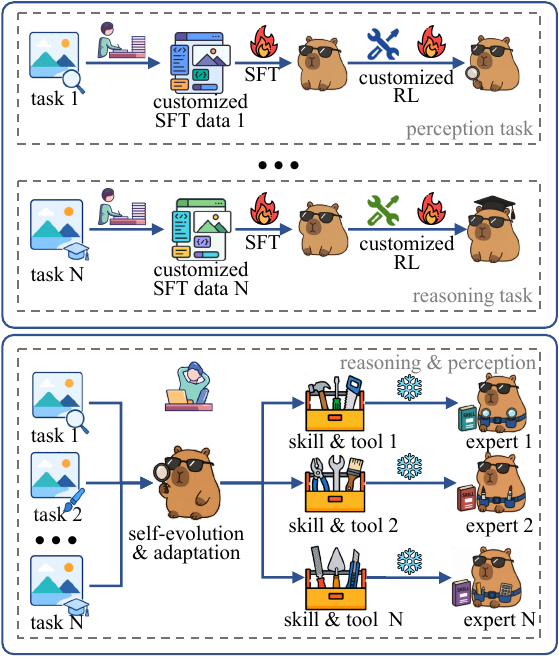}
  \caption{
    \textbf{Hand-crafted SFT/RL vs.\ \sysname{}.}
    Top: each new task family requires hand-curated SFT data and a custom RL pipeline.
    Bottom: \sysname{} instead evolves a task-family-specific skill and tool set from a
    small labeled training subset, with the VLM frozen.
  }
  \label{fig:motivation}
\end{figure}

Vision-language models (VLMs) have improved rapidly, yet adapting a VLM to a new
visual-reasoning task family still typically requires manually curated SFT data or a
hand-designed RL pipeline~\cite{DBLP:conf/cvpr/LiWDLNS25}. We ask whether the agent can build its own task-specific
capability set instead, by inspecting its own behaviour on small training subsets and
without weight updates. Figure~\ref{fig:motivation} contrasts this capability-evolution
route with the typical per-task SFT/RL adaptation pipeline.

We instantiate this idea in \sysname{}, a training-free framework that evolves two
complementary capability types from a small labeled training subset and dynamically
equips them at inference. \textbf{Skills} are
structured Markdown SOPs for \emph{cognitive} bottlenecks, where the visual evidence is
available but the agent's reasoning procedure is weak. \textbf{Tools} are short Python
programs for \emph{perceptual} bottlenecks, where the agent needs a transformed view of
the input before the evidence becomes legible: a crop or zoom into a region of interest,
a contrast adjustment, a chart re-rendering with enlarged axis labels, the extraction of a
specific color or chart layer, or a saliency-guided sub-region extraction. On
each iteration over a sampled sub-training set, \sysname{} diagnoses both the correct and
the incorrect attempts using their images and reasoning traces, proposes multiple candidate
skill$+$tool combinations, and promotes the highest-validation-accuracy candidate into a
persistent library. A mastery phase further learns when each tool is safe to invoke, so
that promoted capabilities are deployed selectively rather than indiscriminately.

We evaluate three questions.
\textbf{(I) Does \sysname{} work across diverse benchmarks and backbones?}
On ChartQA~\citep{DBLP:conf/acl/MasryLTJH22}, MathVista~\citep{DBLP:conf/iclr/LuBX0LH0CG024}, HRBench4K~\citep{DBLP:conf/aaai/Wang0ZZ000T25}, and V$^*$~\citep{DBLP:conf/cvpr/WuX24a} across five VLM backbones, skill-tool co-evolution
improves direct inference on all 20 model--benchmark settings, averaging $+5.6$ accuracy
points.
\textbf{(II) Does the framework extend to a tool-mastery setting with a curated tool set?}
On GTA~\citep{DBLP:conf/nips/WangMLZC0L24}, learning when to call each pre-provided tool
improves step-level tool selection, argument prediction, and instruction-following
accuracy across all tested backbones.
\textbf{(III) Can \sysname{} match task-specific RL across different task families?}
We compare \sysname{} to a representative RL method on two task families: structured
visual QA (VTool-R1~\citep{DBLP:journals/corr/abs-2505-19255}), and
high-resolution perception (DeepEyes~\citep{DBLP:journals/corr/abs-2505-14362}). \sysname{}
matches or approaches RL accuracy at a fraction of the compute (65--99\% gap recovery on
chart/table; 91.1 vs.\ 90.1 on V$^*$-7B perception), and combines additively with RL when
both are available.

\paragraph{Contributions.}
\begin{enumerate}[leftmargin=*, label=(\arabic*), noitemsep]
  \item A \textbf{problem formulation} that recasts per-task VLM adaptation as capability
        evolution: a frozen agent builds its own skill and tool library from a small
        labeled training subset.
  \item A \textbf{multi-candidate evolution loop} that diagnoses correct and incorrect
        attempts, proposes candidate skill$+$tool combinations, promotes the best by
        validation accuracy, learns tools' applicability, and grows the library
        iteratively.
  \item \textbf{Empirical evidence} across four visual-reasoning benchmarks on five VLM
        backbones, the GTA tool-mastery benchmark, and two RL-comparison task families,
        with additive composition when RL is also available.
\end{enumerate}

\section{Related Work}
\label{sec:related_work}

\paragraph{Self-improving agents.}
A growing line of work lets LLM agents improve from experience without gradient updates:
Reflexion~\citep{DBLP:conf/nips/ShinnCGNY23} keeps a verbal reflection buffer that resets
between episodes; Voyager~\citep{DBLP:journals/tmlr/WangX0MXZFA24} and
ExpeL~\citep{DBLP:conf/aaai/Zhao0XLLH24} accumulate persistent skill libraries in game and
tool-use domains; AutoManual~\citep{DBLP:conf/aaai/Zhao0XLLH24} and
EvolveR~\citep{DBLP:journals/corr/abs-2510-16079} refine instruction manuals or reasoning
strategies via self-play; Trace2Skill~\citep{DBLP:journals/corr/abs-2603-25158} distils trajectory-local
lessons into transferable agent skills; and a broader line uses self-distillation to drive
self-evolution~\citep{DBLP:journals/corr/abs-2604-01193,DBLP:conf/acl/SunCZXZY25,DBLP:conf/iclr/XuJNDP0L25}. 


\paragraph{Tool creation for language models.}
A complementary line generates new tools on demand:
CREATOR~\citep{DBLP:conf/emnlp/QianH0Q0J23} prompts an LLM to write Python helpers for problems it
cannot solve directly; LATM~\citep{DBLP:conf/iclr/Cai00CZ24} and ToolLLM~\citep{DBLP:conf/iclr/QinLYZYLLCTQZHT24} scale
this to large API collections. The generated tools are typically text-domain utilities,
such as arithmetic helpers, unit converters, and API wrappers, evaluated on math word
problems and similar text-symbolic benchmarks. 

\paragraph{Visual reasoning with tools.}
A growing line equips VLMs with visual processing tools. VTool-R1~\citep{DBLP:journals/corr/abs-2505-19255},
DeepEyes~\citep{DBLP:journals/corr/abs-2505-14362,DBLP:journals/corr/abs-2511-05271},
PixelReasoner~\citep{DBLP:journals/corr/abs-2505-15966}, and V-Thinker~\citep{DBLP:journals/corr/abs-2511-04460} rely on
SFT and/or RL with curated trajectories to teach VLMs to invoke a fixed tool inventory.
ZoomEye~\citep{DBLP:conf/emnlp/ShenZZXZZY25} and ReFocus~\citep{DBLP:conf/icml/FuLYCLY0FZ25}
are training-free but assume a fixed zoom/search or editing interface;
Lever~LM~\citep{DBLP:conf/nips/YangPMXZHZ24} configures in-context example sequences to
leverage VLMs without retraining. A parallel program-synthesis line composes hand-curated
tool libraries of detectors, OCR, and arithmetic modules
(VisProg~\citep{DBLP:conf/cvpr/GuptaK23}, ViperGPT~\citep{DBLP:conf/iccv/SurisMV23},
Chameleon~\citep{DBLP:conf/nips/LuPCGCWZG23}); MMR-Bench~\citep{DBLP:journals/corr/abs-2601-17814}
evaluates such routing across diverse backbones and tools, and
EcoAlign~\citep{DBLP:journals/corr/abs-2511-11301} frames the cost of adapting VLMs.

\section{Method}
\label{sec:method}

\begin{figure*}[t]
  \centering
  \includegraphics[width=0.97\textwidth,clip]{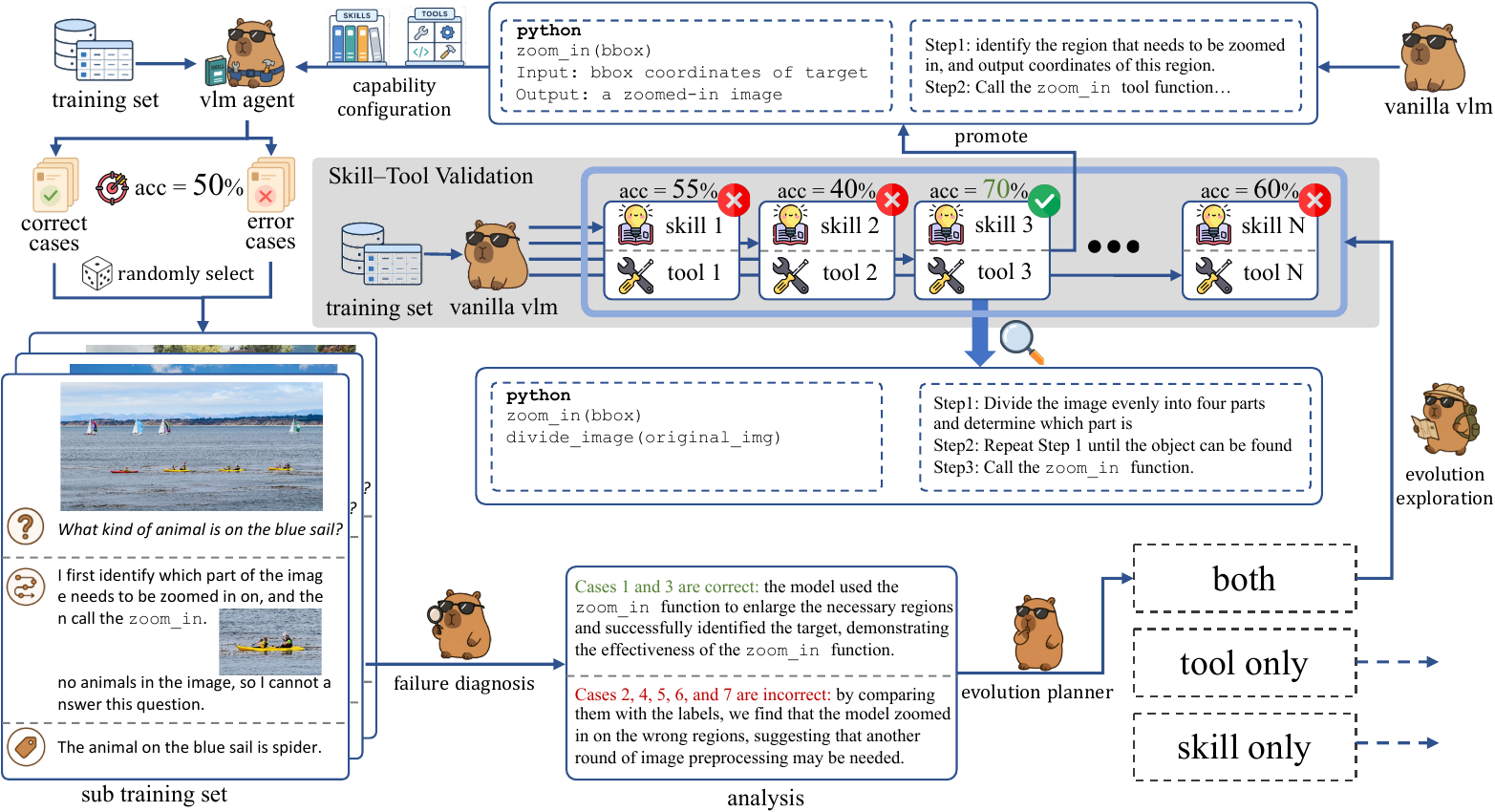}
  \caption{
    \textbf{\sysname{} evolution loop.}
    On a sampled sub-training set, \sysname{} diagnoses both correct and incorrect attempts,
    explores candidate skill$+$tool capabilities, and promotes the best by validation into a
    persistent library.
  }
  \label{fig:method}
\end{figure*}

\subsection{Problem Formulation}
\label{sec:method_formulation}

Let $\mathcal{D}_{\text{train}} = \{(x_i, y_i, \mathbf{v}_i)\}_{i=1}^{k}$ denote a small
training subset of $k$ cases, where $x_i$ is the question, $y_i$ the ground-truth answer,
and $\mathbf{v}_i$ the associated visual input (one or more images). Let $\pi_\theta$ be a
frozen VLM backbone. We define a capability set $\mathcal{C} = (\mathcal{S}, \mathcal{T})$,
where $\mathcal{S}$ is a library of \emph{skills} (structured reasoning SOPs) and
$\mathcal{T}$ is a library of \emph{tools} (executable Python programs). The agent
decomposes solving into a retrieval step and a reasoning step:
\begin{equation}
  f_{\mathcal{C}}(x, \mathbf{v};\, \pi_\theta)
  = \pi_\theta\!\left(\,\cdot \mid x,\, \mathbf{v},\, \mathrm{Retrieve}(x, \mathbf{v};\, \mathcal{C})\right),
  \label{eq:agent}
\end{equation}
where $\mathrm{Retrieve}$ returns the subset of skills and tools relevant to the current
input.
Given a held-out validation set $\mathcal{D}_{\text{val}}$, our goal is to learn
$\mathcal{C}$ that maximises agent accuracy without any weight update:
\begin{equation}
  \mathcal{C}^{\star} = \arg\max_{\mathcal{C}}\; \mathrm{Acc}\!\left(f_{\mathcal{C}};\, \mathcal{D}_{\text{val}}\right),
  \label{eq:objective}
\end{equation}
where the maximisation is over $\mathcal{C}$ only and $\pi_\theta$ remains frozen
($\nabla_\theta = 0$).

\subsection{Evolution Loop}
\label{sec:method_loop}

Each iteration has three
phases (Figure~\ref{fig:method}).

\textbf{Diagnose.}
The agent (Eq.~\ref{eq:agent}) attempts each case in $\mathcal{D}_{\text{train}}$ once with
the current $\mathcal{C}$, producing one reasoning trace $\tau_i$ per case. We then sample
$\mathcal{D}_{\text{sub}} \subseteq \mathcal{D}_{\text{train}}$ across questions (not
stratified per question), including both correct and incorrect attempts when both are
present. If all attempts are correct, the iteration is skipped (no bottleneck to fix); if
all are incorrect, the \textsc{AnalyzerDecider} proceeds on failures alone. The \textsc{AnalyzerDecider} inspects
$\mathcal{D}_{\text{sub}}$---the questions $x_i$, ground-truth answers $y_i$, original
images $\mathbf{v}_i$, reasoning traces $\tau_i$, and any intermediate tool outputs---and
emits a root-cause analysis grounded in visual evidence, an identified bottleneck type (\emph{cognitive} when the visual evidence
is available but the reasoning procedure is weak, or \emph{perceptual} when the agent needs a
transformed visual input), and an action $a \in \{\texttt{skill}, \texttt{tool}, \texttt{both}\}$.
Letting the \textsc{AnalyzerDecider} choose freely from this three-action set on each
case is the default \sysname{} behaviour (denoted \emph{Full} in
Section~\ref{sec:exp1}); restricting $a$ to $\{\texttt{skill}\}$ or $\{\texttt{tool}\}$
yields the \emph{Skill Only} and \emph{Tool Only} ablations.
Visual input here is essential: deciding whether a crop is misaligned, whether labels are
legible, or whether a processed image introduced artefacts requires inspecting the image, not
just the reasoning trace.

\textbf{Explore.}
Conditioned on the diagnosis, the \textsc{Generator} proposes $M$ candidate skill$+$tool
combinations, each addressing the diagnosed bottleneck with a different strategy or
implementation.

\textbf{Validate and promote.}
Writing each candidate as $c_m = (s_m, t_m)$, we evaluate every $c_m$ on
$\mathcal{D}_{\text{train}}$ and promote the highest-accuracy one:
\begin{equation}
\begin{aligned}
  c^{\star} &= \arg\max_{c \in \{c_1, \dots, c_M\}}
              \mathrm{Acc}\!\left(f_{\mathcal{C} \cup c};\, \mathcal{D}_{\text{train}}\right), \\
  \mathcal{C} &\leftarrow
    \begin{cases}
      \mathcal{C} \cup \{c^{\star}\}
        & \text{if } \mathrm{Acc}(f_{\mathcal{C} \cup c^{\star}}) > \mathrm{Acc}(f_{\mathcal{C}}), \\
      \mathcal{C} & \text{otherwise,}
    \end{cases}
\end{aligned}
\label{eq:selection}
\end{equation}
where both accuracies are computed on $\mathcal{D}_{\text{train}}$. The guard ensures
monotonic training-set accuracy: an iteration in which no candidate strictly improves over
the current $f_{\mathcal{C}}$ leaves $\mathcal{C}$ unchanged.
Eq.~\ref{eq:selection} is a training-set proxy for the validation objective in
Eq.~\ref{eq:objective} and subsumes both an \emph{origin} requirement (the new capability
must improve over the previous $f_{\mathcal{C}}$ on the cases it targets) and a
\emph{regression} requirement (it must not degrade currently-correct cases). Because
$\mathcal{D}_{\text{train}}$ and $\mathcal{D}_{\text{val}}$ are disjoint, this proxy does
not contaminate the held-out numbers we report; the only residual concern is selection
bias in $c^{\star}$ over a discrete candidate set, which scales logarithmically in $M$,
so we keep $M$ at a single-digit budget. A broader empirical estimate is left to future
work.
When the action is \texttt{both}, the paired skill additionally serves as the tool's
mastery SOP, gating tool invocation at inference (Section~\ref{sec:method_tool}). The full
procedure runs for $N$ iterations (default $N{=}3$); pseudocode is given in
Appendix~\ref{app:impl}.

\subsection{Skills}
\label{sec:method_skill}

A \emph{skill} is a structured Markdown document with four fields:
\textbf{When-to-Use} (a trigger predicate over question and image type),
\textbf{Strategy} (a numbered step-by-step SOP),
\textbf{Common Pitfalls}, and
\textbf{Worked Example}.
The Generator writes a skill conditioned on the AnalyzerDecider's root-cause analysis; if a
skill covering the same problem class already exists in $\mathcal{S}$, new insights are merged
into it rather than creating a duplicate, preventing library bloat. At inference,
$\mathrm{Retrieve}(\cdot)$ (Eq.~\ref{eq:agent}) returns the top-$K$ skills by BM25
similarity to the question, which the Solver appends to the system prompt.

\subsection{Tools and Mastery}
\label{sec:method_tool}
\label{sec:method_mastery}

A \emph{tool} is a Python function ($\leq$150 lines) that takes one or more image paths and
returns processed images or extracted data, using standard CV libraries (OpenCV, PIL, NumPy).
The Generator writes a tool conditioned on a perceptual bottleneck diagnosis; representative
examples produced in our experiments include chart re-rendering with enlarged axis labels,
saliency-guided sub-region extraction, and contrast enhancement for low-quality documents.

\paragraph{Mastery as a paired skill.}
When the AnalyzerDecider's action is \texttt{both}, the Generator emits a skill alongside the
tool, and this paired skill \emph{is} the tool's mastery SOP. Its When-to-Use predicate
specifies the input patterns on which the tool helps, its Common Pitfalls flag the patterns
on which it hurts, and its Strategy instructs the Solver how to invoke the tool and consume
its output. Because tool invocation is gated by retrieving the paired skill, deployment is
selective by construction rather than indiscriminate---a property that distinguishes
\sysname{} from prior tool-creation methods~\citep{DBLP:conf/emnlp/QianH0Q0J23, DBLP:conf/iclr/Cai00CZ24} that expose
generated tools without learned applicability boundaries.

\paragraph{External tool sets (Mode B).}
When a curated tool set $\mathcal{T}_0 \neq \emptyset$ is provided, \sysname{} can
operate in \textbf{Mode B}: skip tool generation and instead synthesise a mastery skill for
each provided tool $t_j \in \mathcal{T}_0$, learning when to invoke it from the agent's own
behaviour on $\mathcal{D}_{\text{train}}$. Mode B requires no code generation, making it
practical when tool quality is high but deployment strategy is unknown.
Experiment~II (Section~\ref{sec:exp2}) evaluates Mode B.

\subsection{Online Adaptation to Distribution Shift}
\label{sec:method_adaptation}

In real deployments, \sysname{} runs against a streaming query distribution that can shift
over time---a stream may begin with high-resolution perception cases and later start
including MathVista-style reasoning cases, or vice versa. We design \sysname{} to detect
such shifts and \emph{update} its skills and tools online, without weight updates and
without the practitioner having to anticipate the shift or prepare a per-family training
set in advance.

\paragraph{Feedback signal.}
Running the evolution loop online requires a per-case correctness signal to drive
diagnosis and selection (Eqs.~\ref{eq:agent}--\ref{eq:selection}). Raw production traffic
typically lacks ground-truth labels, so \sysname{} assumes one of the practical feedback
channels available in real deployments: a small fraction of queries reviewed by humans in
the loop, an automated quality-assurance pipeline, or an LLM-based verifier that scores
the agent's answer post-hoc. Our experiments use the benchmark's ground-truth labels as a
stand-in for this channel; the adaptation policy itself is agnostic to the source of the
correctness signal, and the framework therefore does not claim fully unsupervised online
adaptation.

\sysname{} continuously samples a sub-training set from the most recent queries (together
with their correctness signals) and runs the evolution loop of
Section~\ref{sec:method_loop} on them, accumulating skills and tools into the library
$\mathcal{C}$. Let $W$ be a rolling-window length. The dominant task family at step $t$ is
\begin{equation}
  \hat{\phi}_t = \arg\max_{\phi}\; \sum_{i = t - W + 1}^{t} \mathbf{1}\!\left[\,\mathrm{fam}(x_i) = \phi\,\right],
  \label{eq:shift}
\end{equation}
where $\mathrm{fam}(\cdot)$ classifies each query from its question and image features.
When Eq.~\ref{eq:shift} flips ($\hat{\phi}_t \neq \hat{\phi}_{t-1}$), the next evolution
iteration is triggered on the new sub-stream and produces updated skills and tools that
handle the new distribution. The library is thus kept in sync with the current query
distribution. We treat this autonomous online adaptation---decoupled from the source of
the correctness signal---as one of the method's contributions; Experiment~I
(Section~\ref{sec:exp1}) evaluates it against a static baseline that never refreshes and
an oracle initialized with the full capability set.

\section{Experiments}
\label{sec:experiments}

\subsection{Experimental Setup}
\label{sec:setup}

\paragraph{Benchmarks.}
For the evolution-from-scratch setting (Exp~I) we use four visual reasoning benchmarks:
\textbf{ChartQA}~\citep{DBLP:conf/acl/MasryLTJH22} (multi-step numerical reasoning over
charts; 2{,}500 test questions),
\textbf{MathVista}~\citep{DBLP:conf/iclr/LuBX0LH0CG024} (math reasoning in figures and
plots; 1{,}000 testmini questions),
\textbf{HRBench4K}~\citep{DBLP:conf/aaai/Wang0ZZ000T25} (perception on 4K-resolution
images; 800 questions), and
\textbf{V$^*$}~\citep{DBLP:conf/cvpr/WuX24a} (visual search for an object's property;
191 questions); the first two stress \emph{cognitive} bottlenecks while the latter two
stress \emph{perceptual} ones, and all four use accuracy as the metric.
For the tool-mastery setting (Exp~II) we use \textbf{GTA}~\citep{DBLP:conf/nips/WangMLZC0L24},
229 real-world tool-use tasks across perception, operation, logic, and creativity
categories, reporting per-step tool-selection, argument-prediction, and
instruction-following accuracy plus end-to-end answer accuracy.
For the RL comparison (Exp~III) we follow the VTool-R1
protocol~\citep{DBLP:journals/corr/abs-2505-19255} on \textbf{ChartQA} and
\textbf{TableQA}, and reuse \textbf{V$^*$} and \textbf{HRBench4K} under the DeepEyes
protocol~\citep{DBLP:journals/corr/abs-2505-14362}.

\paragraph{Foundation models.}
Experiment~I evaluates five proprietary and open-weight VLMs:
\textbf{GPT-4o}~\cite{DBLP:journals/corr/abs-2410-21276}, \textbf{o4-mini}~\cite{openai2025o3o4mini}, \textbf{GPT-5.4}~\cite{openai2026gpt54thinking}, \textbf{Doubao-Seed-2.0}~\cite{bytedanceseed2026seed20}, and
\textbf{Qwen3.5-27B}~\cite{qwen35blog}.
Experiment~II evaluates \textbf{GPT-4o}, \textbf{GPT-5.4}, and \textbf{Doubao-Seed-2.0}
on the GTA benchmark, plus a controlled \textbf{Doubao-Seed-2.0-Pro} analysis that
inspects how provided-tool skills change with tool abstraction and task environment.
Experiment~III uses the \textbf{Qwen2.5-VL} family (3B, 7B) to align with the
VTool-R1 evaluation protocol. All VLM weights are \emph{frozen} throughout; no gradient
updates are performed.

\paragraph{Evolution protocol.}
For each benchmark, we sample 10\% of the official training split as
$\mathcal{D}_{\text{train}}$ and run the \sysname{} evolution loop for $N{=}3$ iterations.
The capability set $\mathcal{C}$ is then frozen and evaluated on the held-out
validation / test split. For each evolution configuration (\emph{Skill Only},
\emph{Tool Only}, \emph{Full}), we run \textbf{three independent seeds} that vary the
10\% training subset and the Generator's sampling, and report mean $\pm$ standard
deviation in Table~\ref{tab:evolution_full}. The \emph{None} (Base Agent) row runs the
frozen VLM with no evolved capabilities once per cell, so no standard deviation is
reported.

\paragraph{Baselines.}
For Experiment~I, we compare four evolution configurations that isolate the contribution
of each component:
\textbf{None (Base Agent)} invokes the frozen VLM with no evolved capabilities (the
zero-shot baseline);
\textbf{Skill Only} restricts the Generator to producing reasoning skills (no tool
generation);
\textbf{Tool Only} restricts the Generator to producing executable visual tools (no skill
generation);
\textbf{Full} is the unrestricted \sysname{} system, where the AnalyzerDecider chooses on
each case whether to generate a skill, a tool, or both.

\paragraph{Evaluation metrics.}
All primary metrics are task-level accuracy on the held-out split.
For Experiment~II we report four of GTA's five canonical metrics: three
step-by-step metrics---\textbf{InstAcc} (Instruction-Following Accuracy: fraction of
steps executed without errors), \textbf{ToolAcc} (Tool Selection Accuracy), and
\textbf{ArgAcc} (Argument Prediction Accuracy)---and the end-to-end \textbf{AnsAcc}
(Answer Accuracy on full tool-using execution). 
For the supplementary controlled GTA analysis, we also report tool usage rate, average
number of tool calls, and dominant tool-call chains.

\subsection{Experiment~I: Autonomous Evolution from Scratch}
\label{sec:exp1}

We test whether \sysname{} can bootstrap a useful capability library from scratch
($\mathcal{S}{=}\emptyset$, $\mathcal{T}{=}\emptyset$) across five backbones and four
benchmarks, and whether the library remains useful when the query distribution shifts
over time (Section~\ref{sec:method_adaptation}).

\paragraph{Per-benchmark results.}
Table~\ref{tab:evolution_full} reports accuracy for the four configurations on each
backbone. Comparing the Full system to the zero-shot \emph{None} baseline, \sysname{}
improves direct inference on all 20 model--benchmark cells, with an average gain of $+5.6$
accuracy points. The largest gains are on o4-mini / V$^*$ ($+14.1$), o4-mini / HRBench4K
($+12.7$), GPT-5.4 / HRBench4K ($+10.7$), Doubao-Seed-2.0 / V$^*$ ($+8.8$), and
Qwen3.5-27B / HRBench4K ($+8.4$). The relative strength of \emph{Skill Only} and \emph{Tool
Only} also follows the cognitive / perceptual split: Skill Only is competitive on ChartQA
and MathVista, where the gain comes from better decomposition and calculation procedures,
while Tool Only and Full dominate on HRBench4K and V$^*$, where sub-region inspection and
visual search are central. The only cell where an ablation edges out Full is Doubao /
HRBench4K, where Tool Only beats Full by $0.0015$---an expected consequence of leaving
capability choice to the AnalyzerDecider rather than enforcing both per case.

\paragraph{Capability quality.}
Appendix~\ref{app:capability_cases} inspects three representative forms of evolved
capability: a ReFocus-style chart/table editing case, a ZoomEye-style coarse-to-fine
search case, and an interleaved skill that pairs a textual SOP with visual evidence. The
case studies illustrate what \sysname{} generates; they do not claim to reproduce the full
algorithms of ReFocus or ZoomEye.

\begin{table*}[t]
\centering
\small
\setlength{\tabcolsep}{5pt}
\renewcommand{\arraystretch}{1.15}
\begin{tabular}{llcccc}
\toprule
\textbf{Model} & \textbf{Evolution Strategy}
  & \textbf{HRBench4K}
  & \textbf{MathVista}
  & \textbf{V$^*$}
  & \textbf{ChartQA} \\
\midrule

\multirow{4}{*}{GPT-4o}
  & None (Base Agent)  &$0.6386$  &$0.6711$  &$0.5828$  &$0.7552$  \\
  & Skill Only         &$\underline{0.6733} {\pm} 0.0121$  &$\underline{0.7122} {\pm} 0.0038$  &$\underline{0.6424} {\pm} 0.0066$  &$\underline{0.7785} {\pm} 0.0026$  \\
  & Tool Only          &$0.6452 {\pm} 0.0044$  &$0.6900 {\pm} 0.0029$  &$0.6225 {\pm} 0.0066$  & $0.7722 {\pm} 0.0029$  \\
  & Full               &$\mathbf{0.6886} {\pm} 0.0071$  &$\mathbf{0.7189} {\pm} 0.0029$  &$\mathbf{0.6689} {\pm} 0.0239$  &$\mathbf{0.7799} {\pm} 0.0045$  \\

\midrule

\multirow{4}{*}{o4-mini}
  & None (Base Agent)  &$0.7300$  &$\underline{0.7533}$  &$0.7285$  &$0.7833$  \\
  & Skill Only         &$\underline{0.7724} {\pm} 0.0058$  &$0.7470 {\pm} 0.0107$  &$0.7616 {\pm} 0.0066$  &$0.7866 {\pm} 0.0151$  \\
  & Tool Only          &$0.7457 {\pm} 0.1670$  &$0.7356 {\pm} 0.0328$  &$\underline{0.8387} {\pm} 0.0629$  & $\underline{0.7920} {\pm} 0.0213$  \\
  & Full               &$\mathbf{0.8571} {\pm} 0.0099$  &$\mathbf{0.7559} {\pm} 0.0055$  &$\mathbf{0.8698} {\pm} 0.0076$  &$\mathbf{0.8007} {\pm} 0.0081$  \\

\midrule

\multirow{4}{*}{GPT-5.4}
  & None (Base Agent)  &$0.7471$  &$0.7589$  &$0.7748$  &$0.7974$  \\
  & Skill Only         &$0.7805 {\pm} 0.0036$  &$\underline{0.7752} {\pm} 0.0039$  &$0.7682 {\pm} 0.0199$  &$0.8090 {\pm} 0.0022$  \\
  & Tool Only          &$\underline{0.8367} {\pm} 0.0346$  &$0.7663 {\pm} 0.0061$  &$\underline{0.8477} {\pm} 0.0132$  & $\underline{0.8108} {\pm} 0.0052$  \\
  & Full               &$\mathbf{0.8538} {\pm} 0.0036$  &$\mathbf{0.7815} {\pm} 0.0101$  &$\mathbf{0.8521} {\pm} 0.0233$  &$\mathbf{0.8276} {\pm} 0.0278$  \\

\midrule

\multirow{4}{*}{Doubao-Seed-2.0}
  & None (Base Agent)  &$0.8214$  &$0.8544$  &$0.8675$  &$0.8152$  \\
  & Skill Only         &$0.8600 {\pm} 0.0049$  &$0.8556{\pm} 0.0109$  &$0.8609 {\pm} 0.0115$  &$\underline{0.8198} {\pm} 0.0031$  \\
  & Tool Only          &$\mathbf{0.9005} {\pm} 0.0022$  &$\underline{0.8659} {\pm} 0.0061$  &$\underline{0.9448} {\pm} 0.0038$  & $0.8156 {\pm} 0.0068$  \\
  & Full               &$\underline{0.8990} {\pm} 0.0064$  &$\mathbf{0.8681} {\pm} 0.0105$  &$\mathbf{0.9558} {\pm} 0.0038$  &$\mathbf{0.8347} {\pm} 0.0295$  \\

\midrule

\multirow{4}{*}{Qwen3.5-27B}
  & None (Base Agent)  &$0.8171$  &$0.8088$  &$0.8808$  &$0.8114$  \\
  & Skill Only         &$0.8721 {\pm} 0.0030$  &$0.8152{\pm} 0.0080$  &$0.8675 {\pm} 0.0066$  &$0.8097 {\pm} 0.0090$  \\
  & Tool Only          &$\underline{0.8986} {\pm} 0.0000$  &$\underline{0.8241} {\pm} 0.0034$  &$\underline{0.9514} {\pm} 0.0101$  & $\underline{0.8115} {\pm} 0.0041$  \\
  & Full               &$\mathbf{0.9010} {\pm} 0.0008$  &$\mathbf{0.8252} {\pm} 0.0023$  &$\mathbf{0.9603} {\pm} 0.0175$  &$\mathbf{0.8158} {\pm} 0.0026$  \\



\bottomrule
\end{tabular}
\caption{%
  \textbf{Autonomous evolution from scratch.} Accuracy across five VLM backbones and four
  benchmarks. The Full \sysname{} configuration improves over the zero-shot \emph{None}
  baseline on all 20 model--benchmark cells.
}
\label{tab:evolution_full}
\end{table*}

\paragraph{Online adaptation to distribution shift.}
To simulate real online user traffic---where the task distribution grows and shifts over
time as different users and requests arrive---we replay a 208-step non-stationary stream
in which each phase introduces a new task family while keeping carry-over cases from
earlier phases: starting with HRBench4K, then adding MathVista, ChartQA, and finally V$^*$
as the new dominant family. The replay reuses completed per-case outputs, and the
benchmark's ground-truth labels stand in for the per-case correctness signal that a
real deployment would obtain from human review, automated QA, or an LLM
verifier~(Section~\ref{sec:method_adaptation}). This isolates the adaptation policy from
the cost of regenerating capabilities online and from the cost of the feedback channel
itself; we do not claim fully unsupervised online adaptation. We compare \emph{Direct} (no capabilities, same as the \emph{None} baseline in
Table~\ref{tab:evolution_full}), \emph{Static} (capabilities evolved on the first phase
only and never refreshed), \emph{Online adapt} (\sysname{} detects the shift via the
rolling-window rule of Section~\ref{sec:method_adaptation} and re-runs the evolution loop
on the new sub-stream), and \emph{Oracle} (pre-evolved on every family; upper bound).

\begin{figure*}[t]
  \centering
  \includegraphics[width=0.95\textwidth]{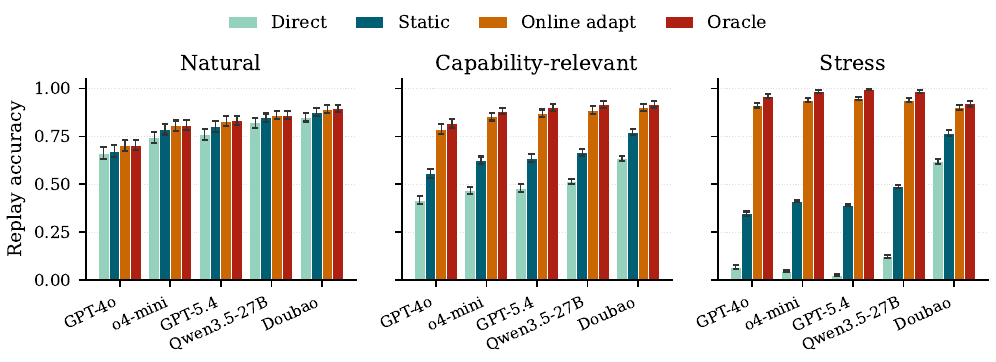}
  \caption{%
    \textbf{Online adaptation to distribution shift.}
    Replay accuracy across five backbones and three stream constructions. \emph{Online
    adapt} stays close to the \emph{Oracle} upper bound and far above \emph{Static}
    everywhere; the gap is largest under the stress protocol, where \emph{Direct} nearly
    collapses. Error bars are seed std (50--200 seeds per cell).
  }
  \label{fig:distribution_shift}
\end{figure*}

Figure~\ref{fig:distribution_shift} compares the four policies across all five backbones
and three stream constructions: the \emph{natural} mixture (a fixed-seed mix of the four
families), \emph{capability-relevant} (cases whose outcome depends on the matching
capability), and \emph{stress} (cases the agent fails on without the matching capability). The
ordering \emph{Direct} $\le$ \emph{Static} $<$ \emph{Online adapt} $\approx$ \emph{Oracle}
is consistent across every backbone $\times$ protocol cell. On GPT-5.4 the gap is
representative: \emph{Online adapt} reaches $0.83$ vs.\ $0.74$ static on natural; widens
to $0.89$ vs.\ $0.65$ on capability-relevant; and on stress, \emph{Direct} collapses to
$0.03$ while \emph{Online adapt} still recovers to $0.95$. The per-step trajectory along
the stream, with shift-detection latencies of 2--7 cases per phase, is shown in
Appendix~\ref{app:distribution_shift_timeseries}.

\subsection{Experiment~II: Learning to Use Pre-Provided Tools}
\label{sec:exp2}

\paragraph{Results on GTA.}
Table~\ref{tab:gta_detailed} compares three methods on the 121-case GTA validation split:
\emph{Base Agent} (zero-shot VLM with the GTA tools exposed); an \emph{XSkill-style}
baseline~\citep{DBLP:journals/corr/abs-2603-12056} that adopts XSkill's prompt-injection
mechanism on top of the same tool environment (see Appendix~\ref{app:xskill_baseline}
for the port); and \sysname{} Mode~B (skills~$+$~mastery learned on the training set).
\sysname{} improves every metric on every backbone, with the largest gains on
Doubao-Seed-2.0 ($+18.7$ ToolAcc, $+15.8$ ArgAcc, $+16.1$ InstAcc, $+4.9$ AnsAcc) and
smaller but uniformly positive gains on GPT-4o ($+4.3$ / $+2.2$ / $+11.4$ / $+5.0$) and
GPT-5.4 ($+6.8$ / $+3.2$ / $+6.1$ / $+1.7$).

\paragraph{What does environment-specific mastery contribute?}
The \emph{XSkill-style} baseline isolates generic skill injection from environment-specific
mastery: it uses the same tools and scorer but its skill is a generic visual-reasoning
prompt without GTA's tool-chain, argument, or answer-normalisation protocols. On every
backbone, \sysname{} beats XSkill-style by $11$--$20$ points on each step-level metric and
$2.5$--$4.9$ on AnsAcc. More tellingly, \emph{XSkill-style itself drops below Base} on the
step-level metrics for GPT-4o and GPT-5.4---adding a generic visual-reasoning prompt
without learning the environment's tool protocol can actually confuse tool selection. The
gain over both baselines is therefore attributable not to skill injection per se, but to
mastery skills that encode the target environment's specific tool chains, argument
formats, and answer normalisation rules.

\begin{table}[t]
\centering
\small
\setlength{\tabcolsep}{4pt}
\renewcommand{\arraystretch}{1.15}
\resizebox{\columnwidth}{!}{%
\begin{tabular}{llcccc}
\toprule
\textbf{Model} & \textbf{Method}
  & \textbf{ToolAcc}
  & \textbf{ArgAcc}
  & \textbf{InstAcc}
  & \textbf{AnsAcc} \\
\midrule

\multirow{3}{*}{GPT-4o}
  & Base Agent      & 80.3 & 62.0 & 52.0 & 66.1 \\
  & XSkill-style    & 71.3 & 53.0 & 49.5 & 68.6 \\
  & \cellcolor{gray!15}\sysname{} (Ours) & \cellcolor{gray!15}\textbf{84.6} & \cellcolor{gray!15}\textbf{64.2} & \cellcolor{gray!15}\textbf{63.4} & \cellcolor{gray!15}\textbf{71.1} \\

\midrule

\multirow{3}{*}{GPT-5.4}
  & Base Agent      & 79.6 & 60.2 & 57.0 & 72.7 \\
  & XSkill-style    & 66.3 & 48.0 & 46.2 & 71.1 \\
  & \cellcolor{gray!15}\sysname{} (Ours) & \cellcolor{gray!15}\textbf{86.4} & \cellcolor{gray!15}\textbf{63.4} & \cellcolor{gray!15}\textbf{63.1} & \cellcolor{gray!15}\textbf{74.4} \\

\midrule

\multirow{3}{*}{Doubao-Seed-2.0}
  & Base Agent      & 67.7 & 45.5 & 41.6 & 76.9 \\
  & XSkill-style    & 72.8 & 48.4 & 44.4 & 76.9 \\
  & \cellcolor{gray!15}\sysname{} (Ours) & \cellcolor{gray!15}\textbf{86.4} & \cellcolor{gray!15}\textbf{61.3} & \cellcolor{gray!15}\textbf{57.7} & \cellcolor{gray!15}\textbf{81.8} \\

\bottomrule
\end{tabular}%
}
\caption{%
  \textbf{GTA results: Base Agent, XSkill-style, and \sysname{} Mode~B on the 121-case
  validation split.} Accuracy (\%) per backbone; best per row in bold.
}
\label{tab:gta_detailed}
\end{table}

\paragraph{Controlled tool-policy analysis.}
Table~\ref{tab:gta_controlled} runs a complementary controlled study on GTA with
Doubao-Seed-2.0-Pro to inspect what the learned policy actually does when either the tool
abstraction or the task environment changes. Two findings stand out. First, fixing the
environment to OCR$+$Calculator, stronger composite tools compress the policy: atomic
tools reach $86.7$ via an \texttt{OCR}$\to$\texttt{Calculator} chain ($22/30$ cases),
while \texttt{VisualArithmeticSolver} reaches $90.0$ with a one-call policy ($30/30$).
Second, the same \texttt{OCR} tool plays different roles in different environments: in
OCR$+$Calc it feeds arithmetic (\texttt{OCR}$\to$\texttt{Calculator}, $19/20$), while in
OCR$+$Search it feeds external lookup
(\texttt{OCR}$\to$\texttt{Search}$\to$\texttt{OCR}$\to$\texttt{Search}, $20/20$).

\begin{table}[t]
\centering
\small
\setlength{\tabcolsep}{4pt}
\renewcommand{\arraystretch}{1.15}
\resizebox{\columnwidth}{!}{%
\begin{tabular}{llcccc}
\toprule
\textbf{Study} & \textbf{Condition} & \textbf{Cases} & \textbf{Acc.}
& \textbf{Tool use} & \textbf{Avg. calls} \\
\midrule
Tool strength & Direct, no tools & 30 & 83.3 & 0.0 & 0.00 \\
Tool strength & Atomic OCR+Calc & 30 & 86.7 & 100.0 & 1.83 \\
Tool strength & Composite VAS & 30 & 90.0 & 100.0 & 1.00 \\
\midrule
OCR role & OCR+Calc env & 20 & 80.0 & 100.0 & 2.20 \\
OCR role & OCR+Search env & 20 & 85.0 & 100.0 & 2.35 \\
\bottomrule
\end{tabular}%
}
\caption{%
  \textbf{Controlled tool-policy analysis on GTA.}
  Acc.\ and Tool use in \%; Avg.\ calls is a count.
}
\label{tab:gta_controlled}
\end{table}


\subsection{Experiment~III: RL and Self-Evolution Across Visual Tool Tasks}
\label{sec:exp3}

We compare \sysname{} against task-specific RL on two task families using two
Qwen2.5-VL backbones (3B, 7B): structured visual QA (ChartQA and TableQA) under
the VTool-R1 protocol~\citep{DBLP:journals/corr/abs-2505-19255}, and high-resolution
perception (V$^\star$ and HRBench4K) under the DeepEyes
protocol~\citep{DBLP:journals/corr/abs-2505-14362}.

\paragraph{ChartQA and TableQA.}
Using the gap-recovery fraction $(\sysname{}-\text{Direct})/(\text{RL}-\text{Direct})$,
\sysname{} recovers $65.7\%$ and $99.4\%$ of the ChartQA gap at 3B and 7B respectively,
and $67.5\%$ and $91.4\%$ of the TableQA gap. \sysname{}\,$+$\,RL further beats RL alone
by $1$--$3$ points on every cell, showing the two interventions compose additively.

\paragraph{High-resolution visual search.}
On V$^\star$, \sysname{} alone beats task-specific RL at both scales ($+6.3$ at 3B,
$+1.0$ at 7B). On HRBench4K, \sysname{} beats RL at 3B by $+4.0$ but trails at 7B by
$1.8$. Adding \sysname{} on top of the RL-trained policy raises 7B V$^\star$ to $92.7$
and 7B HRBench4K to $81.3$, the strongest configurations in those cells.

\paragraph{Cost.}
The RL baselines need on the order of $10^4$--$10^5$ rollouts and $80$--$240$ gradient
updates per backbone (Appendix~\ref{app:rl_cost} for the per-protocol breakdown).
\sysname{} performs no weight updates: a run uses $10\%$ of the training split for
$N{=}3$ passes of frozen-VLM API calls, a few thousand calls per benchmark
(token totals in Appendix~\ref{app:impl}).

\paragraph{Takeaway.}
\sysname{}'s self-evolved skills and tools alone reach RL-level accuracy on both task
families (most of the gap on ChartQA/TableQA; matching or beating RL on V$^\star$ and at
3B on HRBench4K) without any weight updates. RL and self-evolution are not at odds:
\sysname{}\,$+$\,RL is the strongest configuration in every cell, so the two
interventions compose additively.
\begin{table}[t]
\centering
\scriptsize
\setlength{\tabcolsep}{4pt}
\renewcommand{\arraystretch}{1.14}
\resizebox{\columnwidth}{!}{%
\begin{tabular}{llcccc}
\toprule
\textbf{Backbone} & \textbf{Variant}
  & \textbf{ChartQA} & \textbf{TableQA}
  & \textbf{V$^\star$} & \textbf{HRBench4K} \\
\midrule
\multirow{4}{*}{Qwen2.5-VL-3B}
  & Direct              & 45.40 & 43.09 & 71.20 & 63.62 \\
  & RL                  & 66.95 & 56.25 & 78.01 & 66.50 \\
  & \cellcolor{gray!10}\sysname{}      & \cellcolor{gray!10}59.56 & \cellcolor{gray!10}51.97 & \cellcolor{gray!10}84.29 & \cellcolor{gray!10}70.50 \\
  & \cellcolor{gray!25}\sysname{} + RL & \cellcolor{gray!25}\textbf{68.28} & \cellcolor{gray!25}\textbf{59.54} & \cellcolor{gray!25}\textbf{85.34} & \cellcolor{gray!25}\textbf{71.25} \\
\midrule
\multirow{4}{*}{Qwen2.5-VL-7B}
  & Direct              & 56.05 & 56.91 & 79.06 & 70.50 \\
  & RL                  & 75.18 & 68.42 & 84.82 & 76.88 \\
  & \cellcolor{gray!10}\sysname{}      & \cellcolor{gray!10}75.06 & \cellcolor{gray!10}67.43 & \cellcolor{gray!10}85.86 & \cellcolor{gray!10}75.12 \\
  & \cellcolor{gray!25}\sysname{} + RL & \cellcolor{gray!25}\textbf{76.88} & \cellcolor{gray!25}\textbf{69.41} & \cellcolor{gray!25}\textbf{92.67} & \cellcolor{gray!25}\textbf{81.25} \\
\bottomrule
\end{tabular}%
}
\caption{%
  \textbf{Self-evolution vs.\ task-specific RL across two task families.}
  Acc. (\%) on Qwen2.5-VL 3B and 7B.
}
\label{tab:rl_combined}
\end{table}
\subsection{Ablation Studies}
\label{sec:ablation}

The \emph{Skill Only} and \emph{Tool Only} rows of Table~\ref{tab:evolution_full} already
isolate the contribution of each capability type, so we use the remaining ablation budget
to study the \emph{mechanisms} that drive those gains rather than rerunning the same
capability-type comparison. Appendix~\ref{app:ablation_mechanisms} reports three
artifact-backed mechanism probes that support the claims we defend with the current
budget: (i) generic prompt injection does not explain the HRBench4K gain---only the
validated visual-tool path does; (ii) text-only diagnosis cannot materialise a usable
image-processing tool; and (iii) diagnosing only correct or only incorrect ChartQA cases
fails to drive useful evolution, supporting the full loop's use of contrast between
solved and unsolved cases. The remaining mechanism removal (multi-candidate exploration)
needs a larger subset and is left to future work.

\section{Conclusion}
\label{sec:conclusion}

\sysname{} is a training-free framework in which a frozen VLM agent evolves a
task-family-specific library of reasoning skills and executable visual tools from a small
labeled training subset, with the Analyzer choosing on each case whether
to emit a skill, a tool, or both. A mastery skill paired with each generated tool gates
its invocation at inference, and an online adaptation loop refreshes the library when the
query distribution shifts.

Across four visual-reasoning benchmarks and five backbones, \sysname{} improves direct
inference on all 20 model--benchmark cells; on GTA it lifts every step-level tool-use
metric across all tested backbones; and against task-specific RL (VTool-R1, DeepEyes),
\sysname{} matches or beats RL on high-resolution perception, recovers most of the
structured-VQA gap, and composes additively with RL when both are available. These
results suggest much of task-specific VLM adaptation can be obtained without gradient
updates, by letting the agent shape its capability library.

\section*{Limitations}
\label{sec:limitations}

\paragraph{Diagnosis bounded by the backbone.}
The \textsc{AnalyzerDecider}'s bottleneck classification (\emph{cognitive} vs.\
\emph{perceptual}) and the \textsc{Generator}'s candidates are produced by the same frozen
VLM that the agent uses. On backbones with weak self-introspection a mis-classified
bottleneck can propagate to a mis-targeted skill or tool; the promotion guard rejects
under-performing candidates but does not correct upstream diagnosis errors, so smaller or
less capable backbones may converge more slowly or to a less useful library.

\paragraph{Benchmark scope.}
We evaluate capability evolution on image-based VLMs across four visual-reasoning
benchmarks plus GTA and the VTool-R1 / DeepEyes splits. Other modalities
(3D spatial reasoning, medical imaging) are outside the current scope; whether the
cognitive / perceptual decision rule and the executable-tool interface transfer to those
settings is left to future work.

\paragraph{Reliance on a per-case correctness signal.}
Both the offline evolution loop and the online adaptation policy
(Sections~\ref{sec:method_loop}--\ref{sec:method_adaptation}) require a per-case
correctness signal to drive diagnosis and the multi-candidate selection rule. Our
experiments use ground-truth from the benchmark splits as this signal, which
corresponds to deployment settings with human-in-the-loop review, an automated
quality-assurance pipeline, or an LLM-based post-hoc verifier. Deployments that have no
access to any of these channels---i.e.\ fully unsupervised production traffic---fall
outside the scope of the current method; extending \sysname{} to that setting would
require a confidence-based or self-consistency-based feedback substitute that leave
to future work.

\ifqabg
  \bibliographystyle{conference}
\fi
\bibliography{references}

\begin{thebibliography}{36}
\providecommand{\natexlab}[1]{#1}
\providecommand{\url}[1]{\texttt{#1}}
\expandafter\ifx\csname urlstyle\endcsname\relax
  \providecommand{\doi}[1]{doi: #1}\else
  \providecommand{\doi}{doi: \begingroup \urlstyle{rm}\Url}\fi

\bibitem[{ByteDance Seed}(2026)]{bytedanceseed2026seed20}
{ByteDance Seed}.
\newblock Seed2.0 model card: Towards intelligence frontier for real-world
  complexity, February 2026.

\bibitem[Cai et~al.(2024)Cai, Wang, Ma, Chen, and
  Zhou]{DBLP:conf/iclr/Cai00CZ24}
Tianle Cai, Xuezhi Wang, Tengyu Ma, Xinyun Chen, and Denny Zhou.
\newblock Large language models as tool makers.
\newblock In \emph{{ICLR}}. OpenReview.net, 2024.

\bibitem[Cheng et~al.(2025)Cheng, Ma, Ma, and
  Zhang]{DBLP:journals/corr/abs-2511-11301}
Ruoxi Cheng, Haoxuan Ma, Teng Ma, and Hongyi Zhang.
\newblock Ecoalign: An economically rational framework for efficient {LVLM}
  alignment.
\newblock \emph{CoRR}, abs/2511.11301, 2025.

\bibitem[Fu et~al.(2025)Fu, Liu, Yang, Corring, Lu, Yang, Roth,
  Flor{\^{e}}ncio, and Zhang]{DBLP:conf/icml/FuLYCLY0FZ25}
Xingyu Fu, Minqian Liu, Zhengyuan Yang, John Corring, Yijuan Lu, Jianwei Yang,
  Dan Roth, Dinei A.~F. Flor{\^{e}}ncio, and Cha Zhang.
\newblock Refocus: Visual editing as a chain of thought for structured image
  understanding.
\newblock In \emph{{ICML}}, Proceedings of Machine Learning Research. {PMLR} /
  OpenReview.net, 2025.

\bibitem[Gupta \& Kembhavi(2023)Gupta and Kembhavi]{DBLP:conf/cvpr/GuptaK23}
Tanmay Gupta and Aniruddha Kembhavi.
\newblock Visual programming: Compositional visual reasoning without training.
\newblock In \emph{{CVPR}}, pp.\  14953--14962. {IEEE}, 2023.

\bibitem[Hong et~al.(2025)Hong, Zhao, Zhu, Lu, Xu, and
  Yu]{DBLP:journals/corr/abs-2511-05271}
Jack Hong, Chenxiao Zhao, ChengLin Zhu, Weiheng Lu, Guohai Xu, and Xing Yu.
\newblock Deepeyesv2: Toward agentic multimodal model.
\newblock \emph{CoRR}, abs/2511.05271, 2025.
\newblock \doi{10.48550/ARXIV.2511.05271}.
\newblock URL \url{https://doi.org/10.48550/arXiv.2511.05271}.

\bibitem[Jiang et~al.(2026)Jiang, Su, Qu, and
  Fung]{DBLP:journals/corr/abs-2603-12056}
Guanyu Jiang, Zhaochen Su, Xiaoye Qu, and Yi~R. Fung.
\newblock Xskill: Continual learning from experience and skills in multimodal
  agents.
\newblock \emph{CoRR}, abs/2603.12056, 2026.

\bibitem[Li et~al.(2025)Li, Wu, Du, Liu, Nghiem, and
  Shi]{DBLP:conf/cvpr/LiWDLNS25}
Zongxia Li, Xiyang Wu, Hongyang Du, Fuxiao Liu, Huy Nghiem, and Guangyao Shi.
\newblock A survey of state of the art large vision language models: Benchmark
  evaluations and challenges.
\newblock In \emph{{IEEE/CVF} Conference on Computer Vision and Pattern
  Recognition Workshops, {CVPR} Workshops 2025, Nashville, TN, USA, June 11-15,
  2025}, pp.\  1587--1606. Computer Vision Foundation / {IEEE}, 2025.
\newblock URL
  \url{https://openaccess.thecvf.com/content/CVPR2025W/TMM-OpenWorld/html/Li\_A\_Survey\_of\_State\_of\_the\_Art\_Large\_Vision\_Language\_CVPRW\_2025\_paper.html}.

\bibitem[Lu et~al.(2023)Lu, Peng, Cheng, Galley, Chang, Wu, Zhu, and
  Gao]{DBLP:conf/nips/LuPCGCWZG23}
Pan Lu, Baolin Peng, Hao Cheng, Michel Galley, Kai{-}Wei Chang, Ying~Nian Wu,
  Song{-}Chun Zhu, and Jianfeng Gao.
\newblock Chameleon: Plug-and-play compositional reasoning with large language
  models.
\newblock In \emph{NeurIPS}, 2023.

\bibitem[Lu et~al.(2024)Lu, Bansal, Xia, Liu, Li, Hajishirzi, Cheng, Chang,
  Galley, and Gao]{DBLP:conf/iclr/LuBX0LH0CG024}
Pan Lu, Hritik Bansal, Tony Xia, Jiacheng Liu, Chunyuan Li, Hannaneh
  Hajishirzi, Hao Cheng, Kai{-}Wei Chang, Michel Galley, and Jianfeng Gao.
\newblock Mathvista: Evaluating mathematical reasoning of foundation models in
  visual contexts.
\newblock In \emph{{ICLR}}. OpenReview.net, 2024.

\bibitem[Ma et~al.(2026)Ma, Lai, and Ye]{DBLP:journals/corr/abs-2601-17814}
Haoxuan Ma, Guannan Lai, and Han{-}Jia Ye.
\newblock Mmr-bench: {A} comprehensive benchmark for multimodal {LLM} routing.
\newblock \emph{CoRR}, abs/2601.17814, 2026.

\bibitem[Masry et~al.(2022)Masry, Long, Tan, Joty, and
  Hoque]{DBLP:conf/acl/MasryLTJH22}
Ahmed Masry, Do~Xuan Long, Jia~Qing Tan, Shafiq~R. Joty, and Enamul Hoque.
\newblock Chartqa: {A} benchmark for question answering about charts with
  visual and logical reasoning.
\newblock In \emph{{ACL} (Findings)}, Findings of {ACL}, pp.\  2263--2279.
  Association for Computational Linguistics, 2022.

\bibitem[Ni et~al.(2026)Ni, Liu, Liu, Sun, Zhou, Cheng, Wang, Zhao, Jiang, and
  Jiang]{DBLP:journals/corr/abs-2603-25158}
Jingwei Ni, Yihao Liu, Xinpeng Liu, Yutao Sun, Mengyu Zhou, Pengyu Cheng, Dexin
  Wang, Erchao Zhao, Xiaoxi Jiang, and Guanjun Jiang.
\newblock Trace2skill: Distill trajectory-local lessons into transferable agent
  skills.
\newblock \emph{CoRR}, abs/2603.25158, 2026.

\bibitem[OpenAI(2024)]{DBLP:journals/corr/abs-2410-21276}
OpenAI.
\newblock Gpt-4o system card.
\newblock \emph{CoRR}, abs/2410.21276, 2024.

\bibitem[{OpenAI}(2025)]{openai2025o3o4mini}
{OpenAI}.
\newblock Openai o3 and o4-mini system card, April 2025.

\bibitem[{OpenAI}(2026)]{openai2026gpt54thinking}
{OpenAI}.
\newblock Gpt-5.4 thinking system card, March 2026.

\bibitem[Qian et~al.(2023)Qian, Han, Fung, Qin, Liu, and
  Ji]{DBLP:conf/emnlp/QianH0Q0J23}
Cheng Qian, Chi Han, Yi~Ren Fung, Yujia Qin, Zhiyuan Liu, and Heng Ji.
\newblock {CREATOR:} tool creation for disentangling abstract and concrete
  reasoning of large language models.
\newblock In \emph{{EMNLP} (Findings)}, Findings of {ACL}, pp.\  6922--6939.
  Association for Computational Linguistics, 2023.

\bibitem[Qiao et~al.(2025)Qiao, Tan, Yang, Dong, Yang, Lang, Wan, Wang, Xu,
  Yang, Sun, Li, and Zhang]{DBLP:journals/corr/abs-2511-04460}
Runqi Qiao, Qiuna Tan, Minghan Yang, Guanting Dong, Peiqing Yang, Shiqiang
  Lang, Enhui Wan, Xiaowan Wang, Yida Xu, Lan Yang, Chong Sun, Chen Li, and
  Honggang Zhang.
\newblock V-thinker: Interactive thinking with images.
\newblock \emph{CoRR}, abs/2511.04460, 2025.

\bibitem[Qin et~al.(2024)Qin, Liang, Ye, Zhu, Yan, Lu, Lin, Cong, Tang, Qian,
  Zhao, Hong, Tian, Xie, Zhou, Gerstein, Li, Liu, and
  Sun]{DBLP:conf/iclr/QinLYZYLLCTQZHT24}
Yujia Qin, Shihao Liang, Yining Ye, Kunlun Zhu, Lan Yan, Yaxi Lu, Yankai Lin,
  Xin Cong, Xiangru Tang, Bill Qian, Sihan Zhao, Lauren Hong, Runchu Tian,
  Ruobing Xie, Jie Zhou, Mark Gerstein, Dahai Li, Zhiyuan Liu, and Maosong Sun.
\newblock Toolllm: Facilitating large language models to master 16000+
  real-world apis.
\newblock In \emph{{ICLR}}. OpenReview.net, 2024.

\bibitem[Qwen(2026)]{qwen35blog}
Qwen.
\newblock Qwen3.5: Accelerating productivity with native multimodal agents,
  February 2026.
\newblock URL \url{https://qwen.ai/blog?id=qwen3.5}.

\bibitem[Shen et~al.(2025)Shen, Zhao, Zhao, Xu, Zhang, Zhu, and
  Yin]{DBLP:conf/emnlp/ShenZZXZZY25}
Haozhan Shen, Kangjia Zhao, Tiancheng Zhao, Ruochen Xu, Zilun Zhang, Mingwei
  Zhu, and Jianwei Yin.
\newblock Zoomeye: Enhancing multimodal llms with human-like zooming
  capabilities through tree-based image exploration.
\newblock In \emph{{EMNLP}}, pp.\  6602--6618. Association for Computational
  Linguistics, 2025.

\bibitem[Shinn et~al.(2023)Shinn, Cassano, Gopinath, Narasimhan, and
  Yao]{DBLP:conf/nips/ShinnCGNY23}
Noah Shinn, Federico Cassano, Ashwin Gopinath, Karthik Narasimhan, and Shunyu
  Yao.
\newblock Reflexion: language agents with verbal reinforcement learning.
\newblock In \emph{NeurIPS}, 2023.

\bibitem[Su et~al.(2025)Su, Wang, Ren, Lin, and
  Chen]{DBLP:journals/corr/abs-2505-15966}
Alex Su, Haozhe Wang, Weiming Ren, Fangzhen Lin, and Wenhu Chen.
\newblock Pixel reasoner: Incentivizing pixel-space reasoning with
  curiosity-driven reinforcement learning.
\newblock \emph{CoRR}, abs/2505.15966, 2025.

\bibitem[Sun et~al.(2025)Sun, Chen, Zhao, Xu, Zhang, and
  Yin]{DBLP:conf/acl/SunCZXZY25}
Yutao Sun, Mingshuai Chen, Tiancheng Zhao, Ruochen Xu, Zilun Zhang, and Jianwei
  Yin.
\newblock The self-improvement paradox: Can language models bootstrap reasoning
  capabilities without external scaffolding?
\newblock In \emph{{ACL} (Findings)}, Findings of {ACL}, pp.\  6501--6512.
  Association for Computational Linguistics, 2025.

\bibitem[Sur{\'{\i}}s et~al.(2023)Sur{\'{\i}}s, Menon, and
  Vondrick]{DBLP:conf/iccv/SurisMV23}
D{\'{\i}}dac Sur{\'{\i}}s, Sachit Menon, and Carl Vondrick.
\newblock Vipergpt: Visual inference via python execution for reasoning.
\newblock In \emph{{ICCV}}, pp.\  11854--11864. {IEEE}, 2023.

\bibitem[Wang et~al.(2024{\natexlab{a}})Wang, Xie, Jiang, Mandlekar, Xiao, Zhu,
  Fan, and Anandkumar]{DBLP:journals/tmlr/WangX0MXZFA24}
Guanzhi Wang, Yuqi Xie, Yunfan Jiang, Ajay Mandlekar, Chaowei Xiao, Yuke Zhu,
  Linxi Fan, and Anima Anandkumar.
\newblock Voyager: An open-ended embodied agent with large language models.
\newblock \emph{Trans. Mach. Learn. Res.}, 2024, 2024{\natexlab{a}}.

\bibitem[Wang et~al.(2024{\natexlab{b}})Wang, Ma, Li, Zhang, Chen, Chen, and
  Le]{DBLP:conf/nips/WangMLZC0L24}
Jize Wang, Zerun Ma, Yining Li, Songyang Zhang, Cailian Chen, Kai Chen, and
  Xinyi Le.
\newblock {GTA:} {A} benchmark for general tool agents.
\newblock In \emph{NeurIPS}, 2024{\natexlab{b}}.

\bibitem[Wang et~al.(2025)Wang, Ding, Zeng, Zhou, Shen, Luo, Yu, and
  Tao]{DBLP:conf/aaai/Wang0ZZ000T25}
Wenbin Wang, Liang Ding, Minyan Zeng, Xiabin Zhou, Li~Shen, Yong Luo, Wei Yu,
  and Dacheng Tao.
\newblock Divide, conquer and combine: {A} training-free framework for
  high-resolution image perception in multimodal large language models.
\newblock In \emph{{AAAI}}, pp.\  7907--7915. {AAAI} Press, 2025.

\bibitem[Wu et~al.(2025{\natexlab{a}})Wu, Yang, Jiang, Li, Yan, Yu, Zhang,
  Zhai, and Nahrstedt]{DBLP:journals/corr/abs-2505-19255}
Mingyuan Wu, Jingcheng Yang, Jize Jiang, Meitang Li, Kaizhuo Yan, Hanchao Yu,
  Minjia Zhang, Chengxiang Zhai, and Klara Nahrstedt.
\newblock Vtool-r1: Vlms learn to think with images via reinforcement learning
  on multimodal tool use.
\newblock \emph{CoRR}, abs/2505.19255, 2025{\natexlab{a}}.

\bibitem[Wu \& Xie(2024)Wu and Xie]{DBLP:conf/cvpr/WuX24a}
Penghao Wu and Saining Xie.
\newblock V*: Guided visual search as a core mechanism in multimodal llms.
\newblock In \emph{{CVPR}}, pp.\  13084--13094. {IEEE}, 2024.

\bibitem[Wu et~al.(2025{\natexlab{b}})Wu, Wang, Mei, Cai, Fu, Yang, Wen, Yang,
  Shen, Wang, and Shi]{DBLP:journals/corr/abs-2510-16079}
Rong Wu, Xiaoman Wang, Jianbiao Mei, Pinlong Cai, Daocheng Fu, Cheng Yang,
  Licheng Wen, Xuemeng Yang, Yufan Shen, Yuxin Wang, and Botian Shi.
\newblock Evolver: Self-evolving {LLM} agents through an experience-driven
  lifecycle.
\newblock \emph{CoRR}, abs/2510.16079, 2025{\natexlab{b}}.

\bibitem[Xu et~al.(2025)Xu, Jiang, Niu, Deng, Poovendran, Choi, and
  Lin]{DBLP:conf/iclr/XuJNDP0L25}
Zhangchen Xu, Fengqing Jiang, Luyao Niu, Yuntian Deng, Radha Poovendran, Yejin
  Choi, and Bill~Yuchen Lin.
\newblock Magpie: Alignment data synthesis from scratch by prompting aligned
  llms with nothing.
\newblock In \emph{The Thirteenth International Conference on Learning
  Representations, {ICLR} 2025, Singapore, April 24-28, 2025}. OpenReview.net,
  2025.
\newblock URL \url{https://openreview.net/forum?id=Pnk7vMbznK}.

\bibitem[Yang et~al.(2024)Yang, Peng, Ma, Xu, Zhang, Han, and
  Zhang]{DBLP:conf/nips/YangPMXZHZ24}
Xu~Yang, Yingzhe Peng, Haoxuan Ma, Shuo Xu, Chi Zhang, Yucheng Han, and Hanwang
  Zhang.
\newblock Lever {LM:} configuring in-context sequence to lever large vision
  language models.
\newblock In \emph{NeurIPS}, 2024.

\bibitem[Zhang et~al.(2026)Zhang, Bai, Zheng, Jaitly, Collobert, and
  Zhang]{DBLP:journals/corr/abs-2604-01193}
Ruixiang Zhang, Richard~He Bai, Huangjie Zheng, Navdeep Jaitly, Ronan
  Collobert, and Yizhe Zhang.
\newblock Embarrassingly simple self-distillation improves code generation.
\newblock \emph{CoRR}, abs/2604.01193, 2026.

\bibitem[Zhao et~al.(2024)Zhao, Huang, Xu, Lin, Liu, and
  Huang]{DBLP:conf/aaai/Zhao0XLLH24}
Andrew Zhao, Daniel Huang, Quentin Xu, Matthieu Lin, Yong{-}Jin Liu, and Gao
  Huang.
\newblock Expel: {LLM} agents are experiential learners.
\newblock In \emph{{AAAI}}, pp.\  19632--19642. {AAAI} Press, 2024.

\bibitem[Zheng et~al.(2025)Zheng, Yang, Hong, Zhao, Xu, Yang, Shen, and
  Yu]{DBLP:journals/corr/abs-2505-14362}
Ziwei Zheng, Michael Yang, Jack Hong, Chenxiao Zhao, Guohai Xu, Le~Yang, Chao
  Shen, and Xing Yu.
\newblock Deepeyes: Incentivizing "thinking with images" via reinforcement
  learning.
\newblock \emph{CoRR}, abs/2505.14362, 2025.

\end{thebibliography}

\appendix
\clearpage

\section{Implementation Details}
\label{app:impl}

\paragraph{Algorithm.}
Algorithm~\ref{alg:main} gives the full \sysname{} evolution procedure described in
Section~\ref{sec:method_loop}.

\begin{algorithm}[h]
\SetAlgoLined
\KwIn{$\mathcal{D}_{\text{train}}$, frozen VLM $\pi_\theta$, iterations $N$, candidates $M$}
\KwOut{Capability set $\mathcal{C} = (\mathcal{S}, \mathcal{T})$}
$\mathcal{C} \leftarrow (\emptyset, \emptyset)$\;
\For{$n = 1$ \KwTo $N$}{
  Solve $\mathcal{D}_{\text{train}}$ with $f_{\mathcal{C}}$; collect attempts\;
  $\mathcal{D}_{\text{sub}} \leftarrow$ sample from $\mathcal{D}_{\text{train}}$\;
  $a, r \leftarrow \textsc{AnalyzerDecider}(\mathcal{D}_{\text{sub}};\, \pi_\theta)$\;
  $\{(s_m, t_m)\}_{m=1}^{M} \leftarrow \textsc{Generator}(a, r;\, \pi_\theta)$\tcp*{$s_m$ paired with $t_m$ acts as its mastery SOP}
  $m^\star \leftarrow \arg\max_{m} \mathrm{Acc}\bigl(f_{\mathcal{C} \cup (s_m, t_m)};\, \mathcal{D}_{\text{train}}\bigr)$\;
  \If{accuracy improves}{
    $\mathcal{C} \leftarrow \mathcal{C} \cup \{(s_{m^\star}, t_{m^\star})\}$\;
  }
}
\Return{$\mathcal{C}$}\;
\caption{\sysname{} evolution loop.}
\label{alg:main}
\end{algorithm}

\paragraph{Role prompts.}
\sysname{} uses three role-specific system prompts for the same base VLM:
\textit{Solver} (standard QA prompt with capability context injected),
\textit{AnalyzerDecider} (root-cause analysis and decision), and
\textit{Generator} (skill or tool generation).
All prompts are provided in full at \url{[anonymised repository]}.

\paragraph{Skill retrieval.}
At inference time, the Solver retrieves the top-$K{=}3$ skills from $\mathcal{S}$ using BM25
cosine similarity over skill titles and When-to-Use fields.
Retrieved skill text is appended to the system prompt.

\paragraph{Tool invocation.}
The Solver inspects each tool's mastery SOP to decide whether to invoke it on the current input.
If the input matches a tool's supported patterns, the tool is applied and its output image(s)
are appended to the multimodal context.
Tool execution is sandboxed with a 30-second timeout and resource limits.

\paragraph{Hyperparameters.}
Table~\ref{tab:hyperparams} lists all hyperparameter settings.

\begin{table}[h]
\centering
\footnotesize
\setlength{\tabcolsep}{3pt}
\caption{Hyperparameter settings.}
\label{tab:hyperparams}
\begin{tabular}{@{}>{\raggedright\arraybackslash}p{0.28\columnwidth}
                >{\raggedright\arraybackslash}p{0.16\columnwidth}
                >{\raggedright\arraybackslash}p{0.44\columnwidth}@{}}
\toprule
\textbf{Parameter} & \textbf{Value} & \textbf{Description} \\
\midrule
Training subset & 10\% of train split & Per-benchmark evolution subset $\mathcal{D}_{\text{train}}$ \\
$N$               & 3                    & Evolution iterations \\
$K$               & 3                    & Skills retrieved per query \\
Max tool lines    & 150                  & Max lines for generated Python tool \\
Temperature       & 0.7                  & Generator temperature \\
Max tokens (gen.) & 2{,}048              & Max tokens for capability generation \\
\bottomrule
\end{tabular}
\end{table}

\paragraph{Compute.}
Experiments use the frozen VLM backbones listed in Section~\ref{sec:setup}.
Each training case incurs at most three frozen-VLM calls (Solver, AnalyzerDecider, and
Generator), so the per-benchmark call budget scales linearly in the size of the training
subset and in $N$. No GPU training is performed.

\subsection{XSkill-style Baseline}
\label{app:xskill_baseline}

\textbf{What XSkill is.}
XSkill~\citep{DBLP:journals/corr/abs-2603-12056} is a continual-learning framework that
maintains a library of structured Markdown skills (Definition 2.1 of their paper: each
skill carries metadata, a workflow, code templates, and watchpoints) and a complementary
library of short JSON experiences (Definition 2.2: $(c, a, v_e)$ tuples of triggering
condition, recommended action, and semantic embedding, $\leq 64$ words combined). At
inference, XSkill decomposes the task into subtasks, retrieves top-$k$ experiences by
embedding similarity to the current visual context, rewrites them, prunes irrelevant
sections of a retrieved skill against the current image, and injects the adapted skill
plus rewritten experiences into the agent's system prompt as a non-prescriptive reference.

\textbf{Our port to GTA.}
Our XSkill-style baseline keeps XSkill's prompt-injection point in the system prompt, the
same GTA tool environment, and the same scorer as \sysname{} Mode~B. It does not run
XSkill's continual-learning loop end-to-end on GTA training data; instead, the slot that
XSkill would fill with a learned Markdown skill is filled by a single generic
visual-reasoning skill prompt (no GTA-specific tool-chain, argument-format, or
answer-normalisation protocol). This isolates the contribution of XSkill's
skill-injection mechanism from \sysname{}'s GTA-specific mastery skills.

\subsection{RL Baseline Training Compute}
\label{app:rl_cost}

Each RL baseline in Experiment~III is the published, pre-trained checkpoint from its
original protocol, evaluated at both Qwen2.5-VL scales (3B, 7B).

\paragraph{VTool-R1 (ChartQA, TableQA).}
The released checkpoints are reported after $80$ GRPO steps with a batch of $32$ prompts
and $n_{\text{rollout}}{=}8$ rollouts per prompt, giving approximately
$80 \times 32 \times 8 = 20{,}480$ trajectories and $80$ gradient updates per backbone.

\paragraph{DeepEyes (V$^\star$, HRBench4K).}
The released checkpoints are reported after $240$ RL steps with a batch of $128$ prompts
and $n_{\text{rollout}}{=}4$ rollouts per prompt, giving approximately
$240 \times 128 \times 4 = 122{,}880$ trajectories and $240$ gradient updates per backbone.

\paragraph{\sysname{}.}
A \sysname{} run performs no gradient updates. It uses $10\%$ of each benchmark's training
split as $\mathcal{D}_{\text{train}}$ and runs the evolution loop for $N{=}3$ passes,
invoking the frozen VLM only for solving, diagnosis, generation, and validation. The
total budget is on the order of a few thousand frozen-VLM API calls per benchmark;
per-benchmark token totals appear earlier in this appendix.

\section{Mechanism Ablations}
\label{app:ablation_mechanisms}

Experiment~I in the main paper already isolates the contribution of each capability
\emph{type}: the \emph{None}, \emph{Skill Only}, \emph{Tool Only}, and \emph{Full} rows of
Table~\ref{tab:evolution_full} answer whether skills, tools, or both account for the
observed gains. The ablations here test a complementary question: when the Full system is
used, which \emph{mechanisms} inside the evolution loop (Section~\ref{sec:method_loop}) are
doing the work? Each variant removes one design decision and holds the rest fixed, so each
row of Table~\ref{tab:ablation_mechanism} rules out one alternative explanation for the
Full system's gains.

\paragraph{Setup.}
The full grid of mechanism ablations is expensive because each row rebuilds
$\mathcal{C}$ from scratch. We therefore report the completed artifact-backed probes first,
and use them to answer the most immediate reviewer question: is the gain merely generic
prompting, or does the evolved visual-tool path matter? All rows in
Table~\ref{tab:ablation_mechanism} use the same HRBench4K validation set (700 cases) and the
same function-calling VQA runtime with Doubao-Seed-2.0-Pro. For the first three rows,
tools are disabled, so the comparison isolates prompt/skill content; the last two rows
reuse the existing full-val tool-enabled runs for context.

\paragraph{Completed mechanism probe.}
Table~\ref{tab:ablation_mechanism} shows that generic skill injection does not explain
the HRBench4K gain. A neutral task prompt is slightly below the no-capability runtime, and
the evolved no-tool skill is essentially tied with it. In contrast, enabling the evolved
visual-tool/artifact path yields a much larger improvement, and pairing the evolved tool
with its mastery skill (\emph{Evolved Skill + Tool}) reaches the highest accuracy in the
probe. This supports the mechanism claim for perceptual benchmarks: the active ingredient
is not ``more prompt text'' but validated image transformations that expose hard-to-read
local evidence, with paired mastery skills adding a small incremental boost on top.

\begin{table*}[t]
\centering
\small
\setlength{\tabcolsep}{4pt}
\renewcommand{\arraystretch}{1.18}
\begin{tabular}{lp{0.38\textwidth}ccc}
\toprule
\textbf{Variant} & \textbf{Mechanism isolated}
  & \textbf{Tools} & \textbf{Correct / Total} & \textbf{Accuracy} \\
\midrule
No Capability                & no skill context; no tool exposure
  & off & 599 / 700 & 0.8557 \\
Generic Skill                & evolved library $\to$ hand-neutral HRBench4K prompt
  & off & 594 / 700 & 0.8486 \\
Evolved No-Tool Skill        & evolved HRBench4K SOP, but visual tools disabled
  & off & 597 / 700 & 0.8529 \\
\midrule
Evolved Tool Only            & validated visual tools/artifacts, no extra generated skill text
  & on & 625 / 700 & 0.8929 \\
Evolved Skill + Tool         & paired evolved skill/tool deployment
  & on & 629 / 700 & 0.8986 \\
\bottomrule
\end{tabular}
\caption{%
  \textbf{Completed HRBench4K mechanism probe.}
  The first three rows disable tools and test whether generic prompt/skill text explains
  the gain. It does not: generic and evolved no-tool skills are near the no-capability
  runtime. The tool-enabled rows show a much larger gain from validated visual artifacts,
  and the paired \emph{Evolved Skill + Tool} reaches the highest accuracy in the probe,
  isolating both the perceptual mechanism and the additional boost from pairing skill
  with tool.
}
\label{tab:ablation_mechanism}
\end{table*}

\paragraph{Guarded 30-case tool-generation pilot.}
After adding the generated-tool acceptance guards, we ran a time-bounded HRBench4K pilot on
the first 30 validation cases to check whether the mechanism interventions now instantiate
the intended tool path. This pilot is not a replacement for the 700-case mechanism probe
above; it is an audit of the evolution loop under the stricter acceptance criterion. We
therefore report the selection score used by the loop---candidate accuracy on the same
30-case subset against the same-run baseline---rather than treating the noisy final replay
as a held-out estimate.

\begin{table*}[t]
\centering
\small
\setlength{\tabcolsep}{4pt}
\renewcommand{\arraystretch}{1.15}
\begin{tabular}{lp{0.42\textwidth}ccc}
\toprule
\textbf{Variant} & \textbf{Guarded outcome}
  & \textbf{Baseline} & \textbf{Candidate} & $\Delta$ \\
\midrule
Full tool-guard pilot & accepted one generated tool with its paired mastery skill;
target HRBench4K-cross family improved from 0.5000 to 0.8125
  & 0.5667 & 0.6000 & +0.0333 \\
Text-only Diagnosis & rejected: the proposed tool produced no visual artifact, so no
generated tool was promoted
  & 0.6667 & 0.6667 & 0.0000 \\
No Paired Mastery & smoke passed, but the unpaired tool candidate regressed on the full
subset; HRBench4K-cross fell from 0.5625 to 0.4375
  & 0.5667 & 0.5333 & -0.0333 \\
No Persistence & the candidate regressed before persistence mattered, so this row is
diagnostic but not a clean persistence ablation
  & 0.5667 & 0.5333 & -0.0333 \\
\bottomrule
\end{tabular}
\caption{%
  \textbf{Guarded HRBench4K tool-generation pilot on 30 validation cases.}
  All rows require at least one generated tool and forbid skill-only fallback. Numbers
  are selection scores against the same-run baseline (not held-out effect sizes), so the
  rows should be read as a mechanism audit rather than a benchmark comparison.
}
\label{tab:ablation_guarded_pilot}
\end{table*}

\subsection{Generic Skill vs. Evolved Capability}
\label{app:abl_generic}

\paragraph{What is removed.}
The evolved capability library is replaced with a single hand-neutral task prompt that
gives generic task advice (for ChartQA: ``read the chart carefully, identify relevant
values, reason step by step''; for HRBench4K: ``inspect local details before answering and
verify the target object''). No diagnosis, generation, validation, or persistence is run.

\paragraph{What this rules out.}
A pure prompt-engineering account would predict that Generic Skill closes most of the gap
to Full, since the evolved skills are themselves natural-language SOPs. If Generic Skill
improves over Direct by only a small margin and stays well below Full, the gap is
attributable to mechanism-level evolution---failure-specific procedures, applicability
triggers, and pitfalls that a generic prompt cannot anticipate---rather than to the
presence of any task-aware prompt.

\paragraph{Observed pattern.}
On HRBench4K with tools disabled, Generic Skill obtains 0.8486 accuracy, compared with
0.8557 for the no-capability runtime and 0.8529 for the evolved no-tool skill. These rows
show that simply adding a neutral task prompt does not recover the tool-enabled gains.
The tool-enabled rows in Table~\ref{tab:ablation_mechanism} are roughly four points
higher, so the HRBench4K improvement is better explained by validated visual artefacts than
by generic skill injection.

\subsection{Correct-Only vs. Error-Only Diagnosis}
\label{app:abl_diagnosis_polarity}

\paragraph{What is removed.}
This stress test splits the diagnosis evidence by polarity on a ChartQA $k{=}50$,
$N{=}3$ run. Correct-Only Diagnosis exposes only cases the current agent already solves;
Error-Only Diagnosis exposes only the cases it misses. All other settings match the
ChartQA structured evolution protocol.

\paragraph{What this rules out.}
The full method assumes that useful evolution needs contrast: correct cases show the
boundary of the existing capability, while incorrect cases expose the missing behavior.
Correct-only evidence should therefore have no target for evolution, and error-only
evidence should over-focus on the visible failures without enough successful neighboring
cases to constrain the proposed capability.

\begin{table*}[t]
\centering
\small
\setlength{\tabcolsep}{4pt}
\renewcommand{\arraystretch}{1.14}
\begin{tabular}{lp{0.38\textwidth}ccc}
\toprule
\textbf{Variant} & \textbf{Observed evolution behavior}
  & \textbf{Rounds} & \textbf{Promoted} & \textbf{Train replay} \\
\midrule
Correct-Only Diagnosis & no failure cluster is constructed, so the loop stops before
candidate generation
  & 0 & 0 & 45 / 50 \\
Error-Only Diagnosis & each round targets the same ChartQA error cluster and proposes a
new broad visual-overlay tool; round 1 fails smoke validation, while rounds 2--3 tie the
baseline selection score and are rejected
  & 3 & 0 & 45 / 50 \\
\bottomrule
\end{tabular}
\caption{%
  \textbf{Diagnosis-polarity stress test on ChartQA.}
  Correct-only evidence cannot drive evolution because no missed cases are available.
  Error-only evidence repeatedly triggers new tool generation
  (\texttt{chart\_data\_point\_highlighter},
  \texttt{chart\_series\_value\_overlay}, and
  \texttt{chart\_data\_point\_overlay\_generator}) but promotes none of them. The
  selection baseline and candidates for the two completed error-only rounds are both
  46/50, so the generated tools do not improve the subset despite repeated regeneration.
}
\label{tab:ablation_diagnosis_polarity}
\end{table*}

\paragraph{Observed pattern.}
The two extremes show the intended qualitative failure modes. Correct-only diagnosis is
degenerate: with no failures in the digest, the evolution loop has no target family to
repair and accepts no capability. Error-only diagnosis is active but unstable: the decider
asks for \texttt{generate\_both} in every round, and the generator produces a fresh
overlay/highlighter-style tool each time rather than refining a stable reusable
capability. Since none of these candidates improves over the 46/50 selection baseline,
the library remains unchanged. The result supports the design choice to diagnose both
correct and incorrect attempts rather than treating either polarity alone as sufficient.

\subsection{Text-only Diagnosis}
\label{app:abl_textonly}

\paragraph{What is removed.}
The AnalyzerDecider's inputs are restricted to text only: the question, the agent's answer
(correct or incorrect), the ground-truth answer, and the reasoning trace. No original
image, no intermediate tool output, and no processed visual artefact is passed in.

\paragraph{What this rules out.}
Text reflection can say \emph{that} an answer was wrong, but it cannot tell whether a
crop was misaligned, whether a zoomed region missed the target, or whether a processed
image introduced artefacts. We expect the gap to Full to be largest on the perceptual
benchmarks (HRBench4K, V$^\star$). On ChartQA the gap should be smaller because many
failures there are procedural; a small ChartQA gap with a large HRBench4K/V$^\star$ gap is
itself the signature of the visual-diagnosis claim.

\paragraph{Observed pattern.}
The guarded HRBench4K pilot produced the expected failure mode: the text-only candidate was
rejected because the proposed tool did not produce a visual artifact. Its selection score
therefore stayed at the same-run baseline (0.6667 to 0.6667) and no generated tool was
promoted. This supports the qualitative mechanism claim that visual diagnosis is needed to
materialize useful image-processing tools, not just to write another textual rule.

\subsection{No Paired Mastery Skill}
\label{app:abl_mastery}

\paragraph{What is removed.}
When the AnalyzerDecider's action is \texttt{both}, the Generator emits only the tool and
not its paired mastery skill. The tool is added to $\mathcal{T}$ but has no
$\textsc{When-to-Use}$ predicate gating its invocation at inference, so any skill
retrieval that surfaces a tool reference exposes the tool unconditionally.

\paragraph{What this rules out.}
Prior tool-creation methods deploy generated tools indiscriminately. The paired-skill
design in Section~\ref{sec:method_mastery} predicts that gating tool invocation by
retrieval of an applicability-encoding skill is what keeps tool usage selective.
No Paired Mastery Skill is expected to keep or even increase tool-usage rate but reduce
accuracy on benchmarks where naive tool exposure can hurt---MathVista is the canonical
cautionary case.

\paragraph{Observed pattern.}
The guarded HRBench4K pilot cleanly instantiated this ablation. The unpaired generated tool
passed smoke validation, but once exposed without a mastery skill it regressed on the
30-case subset (0.5667 to 0.5333) and was rejected. The targeted HRBench4K-cross family also
dropped from 0.5625 to 0.4375. This is the strongest new mechanism evidence from the
pilot: the tool itself can be executable and locally plausible, while the missing
when-to-use gate still makes it harmful at subset scale.

\subsection{No Persistence}
\label{app:abl_persistence}

\paragraph{What is removed.}
Capabilities promoted during one iteration are not stored in $\mathcal{C}$. The agent can
still reflect on a case and immediately retry it with the generated capability, but the
capability is discarded before the next case is processed, so no cross-case retrieval
occurs.

\paragraph{What this rules out.}
A retry-only account would predict that most of Full's gain comes from in-place retry on
the triggering case rather than from cross-case reuse. If No Persistence matches Full on
held-out accuracy, the library is a bookkeeping device with no inferential value. A gap
shows that the system's advantage comes from accumulating capabilities that future cases
can retrieve, not just from one more attempt at the triggering case. ChartQA is the
cleanest test because failures there form recurring families (stacked-bar boundary
reading, multi-series selection) that a single capability can fix across many held-out
cases.

\paragraph{Observed pattern.}
In the guarded HRBench4K pilot, the generated candidate regressed on the subset
(0.5667 to 0.5333) and was rejected by ordinary candidate selection before the
persistence switch could be exercised; we therefore mark this row as inconclusive for the
cross-case persistence claim on this particular pilot.

\subsection{Summary}
\label{app:abl_summary}

We deliberately keep the claims from these tables narrow.

\textbf{Generic prompt injection does not explain the HRBench4K gain.}
Table~\ref{tab:ablation_mechanism} (700 cases) shows that disabling the tool path and
replacing the evolved library with either a hand-neutral prompt or an evolved no-tool
skill leaves accuracy within noise of the no-capability runtime; the four- to five-point
gain appears only once the validated visual-tool path is enabled. This rules out a pure
prompt-engineering account of \sysname{} on perceptual tasks.

\textbf{Visual diagnosis is necessary to materialise a usable image-processing tool.}
In the guarded 30-case pilot (Table~\ref{tab:ablation_guarded_pilot}), the text-only
diagnosis variant proposed a tool that produced no visual artifact and was rejected by
the acceptance guard, leaving the selection score equal to the same-run baseline.

\textbf{Diagnosis must see both correct and incorrect cases.}
The ChartQA polarity stress test in Table~\ref{tab:ablation_diagnosis_polarity} shows that
neither polarity alone supports useful evolution: correct-only diagnosis cannot enter the
generation path because no failure cluster is formed, while error-only diagnosis is
active but unstable, regenerating a fresh broad overlay tool in each round and promoting
none of them. The contrast between solved and unsolved cases is what gives the
AnalyzerDecider a stable target for capability proposal.

The remaining two pilot rows (No Paired Mastery and No Persistence) sit within
subset-level noise on 30 cases and should be read as diagnostic audits of the loop
machinery rather than as effect-size estimates. The multi-candidate exploration
mechanism (single-candidate vs.\ $M{>}1$) requires a larger-subset experiment than the
current pilot allows and is left to future work.

\paragraph{Practical diagnostic.}
The mechanism ablations also suggest a practical rule of thumb for new task families. If
the bottleneck is procedural---visual evidence is present but the reasoning is weak---skill
evolution is the lower-cost first choice; if a different view of the image is required
before the evidence becomes accessible, tool evolution is necessary. A practitioner
targeting a new benchmark can use a small pilot run (e.g.\ 5\% of the training split with
$N{=}1$) to inspect which kinds of capabilities are generated and whether they transfer
beyond the triggering examples.

\section{Additional Results}
\label{app:results}

\subsection{Training Set Size Sensitivity}

\begin{figure}[ht]
\centering
\includegraphics[width=0.82\columnwidth]{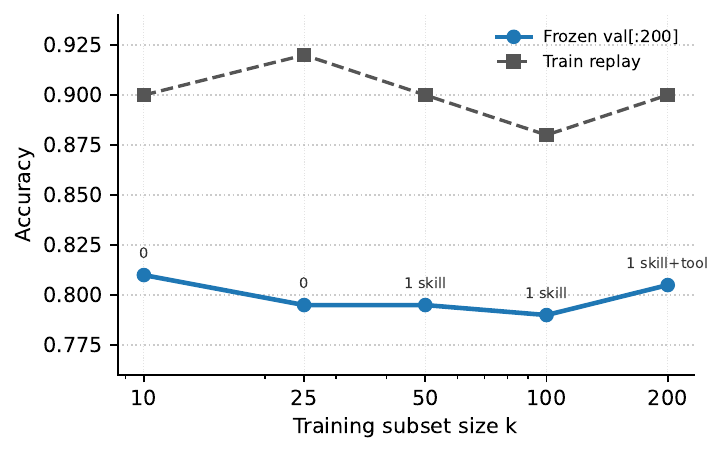}
\caption{%
  \textbf{ChartQA training-size sensitivity.}
  We sweep $k \in \{10,25,50,100,200\}$ with $N{=}3$ evolution iterations and
  evaluate each frozen library on the same 200-case ChartQA validation slice.
  Labels above the held-out curve indicate the promoted capabilities.
}
\label{fig:k_sensitivity}
\end{figure}

\begin{table}[ht]
\centering
\small
\setlength{\tabcolsep}{4pt}
\renewcommand{\arraystretch}{1.12}
\resizebox{\columnwidth}{!}{%
\begin{tabular}{rccc}
\toprule
\textbf{$k$} & \textbf{Promoted capabilities} & \textbf{Train replay} & \textbf{Frozen val[:200]} \\
\midrule
10  & 0 & 9 / 10   (0.900) & 162 / 200 (0.810) \\
25  & 0 & 23 / 25  (0.920) & 159 / 200 (0.795) \\
50  & 1 skill & 45 / 50  (0.900) & 159 / 200 (0.795) \\
100 & 1 skill & 88 / 100 (0.880) & 158 / 200 (0.790) \\
200 & 1 skill + 1 tool & 180 / 200 (0.900) & 161 / 200 (0.805) \\
\bottomrule
\end{tabular}}
\caption{%
  \textbf{Artifact-backed ChartQA sensitivity sweep.}
  All rows use the same safe ChartQA normalization, Doubao-Seed-2.0-Pro runtime,
  $N{=}3$, max-attempts 5, and the same 200-case validation slice.
}
\label{tab:k_sensitivity}
\end{table}

Figure~\ref{fig:k_sensitivity} and Table~\ref{tab:k_sensitivity} show that the held-out
accuracy is stable across training subset sizes, ranging from 0.790 to 0.810 on the
matched validation slice. The sweep therefore supports a saturation interpretation rather
than a tuning claim: small subsets already expose the common ChartQA reading patterns,
while increasing $k$ mainly changes what the gate is willing to promote. In particular,
the $k{=}200$ run is the only row that promotes a generated visual tool together with its
mastery skill, and it recovers the best held-out score among the evolved-library rows
(0.805), but the difference from smaller subsets remains within two accuracy points.

\subsection{Distribution-Shift Adaptation: Time Series and Aggregates}
\label{app:distribution_shift_timeseries}

\begin{figure}[ht]
  \centering
  \includegraphics[width=\columnwidth]{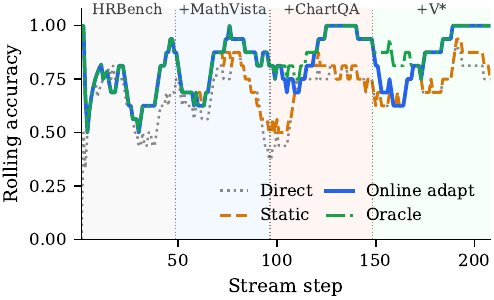}
  \caption{%
    \textbf{Rolling accuracy along the natural stream (GPT-5.4).}
    Phase labels above the plot mark the newly introduced family at each shift; earlier
    families remain present in the mix. \emph{Static} lags at each shift; \emph{Online
    adapt} catches up to \emph{Oracle} within a 2--7 case detection latency.
  }
  \label{fig:distribution_shift_timeseries}
\end{figure}

Figure~\ref{fig:distribution_shift} in the main paper aggregates the distribution-shift
experiment across backbones and stream constructions. This appendix shows the per-step
trajectory along the stream (Figure~\ref{fig:distribution_shift_timeseries}) and the
full per-backbone numerical aggregates (Table~\ref{tab:distribution_shift}).

\begin{table}[ht]
\centering
\small
\setlength{\tabcolsep}{4pt}
\renewcommand{\arraystretch}{1.12}
\resizebox{\columnwidth}{!}{%
\begin{tabular}{lcccc}
\toprule
\textbf{Replay protocol}
  & \textbf{Direct}
  & \textbf{Static}
  & \textbf{Online adapt}
  & \textbf{Oracle} \\
\midrule
\multicolumn{5}{l}{\emph{GPT-4o (Full)}} \\
Capability-relevant & $0.418{\pm}0.020$ & $0.558{\pm}0.023$ & $0.789{\pm}0.025$ & $0.818{\pm}0.022$ \\
Stress & $0.068{\pm}0.009$ & $0.347{\pm}0.011$ & $0.912{\pm}0.008$ & $0.961{\pm} 0.010$ \\
Natural & $0.664{\pm}0.032$ & $0.673{\pm}0.030$ & $0.702{\pm}0.028$ & $0.705{\pm}0.028$ \\
\midrule
\multicolumn{5}{l}{\emph{o4mini (Full)}} \\
Capability-relevant & $0.467{\pm}0.019$ & $0.626{\pm}0.019$ & $0.852{\pm}0.019$ & $0.881{\pm}0.016$ \\
Stress & $0.047{\pm}0.006$ & $0.412{\pm}0.006$ & $ 0.940{\pm}0.009$ & $0.984{\pm}0.007$ \\
Natural & $0.745{\pm}0.029$ & $0.787{\pm}0.028$ & $0.807{\pm}0.027$ & $0.808{\pm}0.027$ \\
\midrule
\multicolumn{5}{l}{\emph{GPT-5.4 (Full)}} \\
Capability-relevant & $0.479{\pm}0.021$ & $0.636{\pm}0.020$ & $0.870{\pm}0.020$ & $0.899{\pm}0.018$ \\
Stress & $0.026{\pm}0.004$ & $0.391{\pm}0.004$ & $ 0.947{\pm}0.007$ & $0.995{\pm}0.005$ \\
Natural & $0.761{\pm}0.029$ & $0.800{\pm}0.029$ & $0.830{\pm}0.026$ & $0.833{\pm}0.025$ \\
\midrule
\multicolumn{5}{l}{\emph{Qwen3.5-27B (Full)}} \\
Capability-relevant & $0.514{\pm}0.015$ & $0.667{\pm}0.016$ & $ 0.888{\pm}0.020$ & $ 0.917{\pm}0.018$ \\
Stress & $0.122{\pm}0.007$ & $0.488{\pm}0.007$ & $0.940{\pm}0.009$ & $0.984{\pm}0.008$ \\
Natural & $0.821{\pm}0.026$ & $0.847{\pm}0.023$ & $0.861{\pm}0.022$ & $0.862{\pm}0.022$ \\
\midrule
\multicolumn{5}{l}{\emph{Doubao-Seed-2.0 (Full)}} \\
Capability-relevant & $0.635{\pm}0.014$ & $0.773{\pm}0.015$ & $0.900{\pm}0.018$ & $ 0.917{\pm}0.016$ \\
Stress & $0.621{\pm}0.013$ & $0.768{\pm}0.014$ & $0.902{\pm}0.014$ & $0.920{\pm}0.014$ \\
Natural & $0.850{\pm}0.023$ & $0.876{\pm}0.022$ & $0.893{\pm}0.019$ & $0.895{\pm}0.019$ \\
\bottomrule
\end{tabular}%
}
\caption{%
  \textbf{Per-backbone aggregates for the distribution-shift experiment.}
  Mean accuracy with seed std (50--200 seeds per cell). Bars in
  Figure~\ref{fig:distribution_shift} are computed from this table.
}
\label{tab:distribution_shift}
\end{table}

\subsection{Multi-benchmark Joint Evolution}
\label{app:mixed_benchmark}

Experiment~I in the main paper evolves a separate library per benchmark. This appendix
tests whether a single shared library can serve all four visual reasoning benchmarks at
once. We combine the validation sets of \textbf{ChartQA}, \textbf{MathVista},
\textbf{HRBench4K}, and \textbf{V$^\star$} into one mixed pool, sample 10\% as the joint
evolution subset, and use the remaining 3{,}671 cases as the held-out test set. The
evolution loop produces a single unified skill paired with the standard preset tools
(\texttt{execute\_python}, \texttt{get\_image\_info}, \texttt{zoom\_image},
\texttt{crop\_image}); we evaluate the frozen library on the held-out pool with GPT-4o.

\begin{table}[ht]
\centering
\small
\setlength{\tabcolsep}{4pt}
\renewcommand{\arraystretch}{1.12}
\begin{tabular}{lccc}
\toprule
\textbf{Benchmark} & \textbf{Held-out cases} & \textbf{Correct} & \textbf{Accuracy} \\
\midrule
ChartQA      & 1{,}920 & 1{,}488 & 0.775 \\
MathVista    &    900 &     616 & 0.684 \\
HRBench4K      &    700 &     476 & 0.680 \\
V$^\star$    &    151 &     101 & 0.669 \\
\midrule
\textbf{Combined} & \textbf{3{,}671} & \textbf{2{,}681} & \textbf{0.730} \\
\bottomrule
\end{tabular}
\caption{%
  \textbf{Multi-benchmark joint evolution (GPT-4o).}
  A single unified skill, evolved on a 10\% subset of the combined four-benchmark
  validation pool, is applied to the remaining 3{,}671 held-out cases. The same library
  is used across all four benchmarks; no per-benchmark capability is loaded.
}
\label{tab:mixed_benchmark}
\end{table}

The unified skill reaches $73.0\%$ overall on the held-out pool, with per-benchmark
accuracy ranging from $66.9\%$ on V$^\star$ to $77.5\%$ on ChartQA. This shows that
\sysname{} can evolve a library that generalises across visually distinct benchmark
families from a single joint training subset, rather than requiring a separate evolution
run per benchmark---an option useful in practice when a deployed agent must answer
queries drawn from a mixture of task families.

\section{Qualitative Analysis of Evolved Capabilities}
\label{app:capability_cases}

This appendix gives three qualitative case studies of evolved capabilities. The examples are
intended to make the generated artifacts concrete, not to introduce new quantitative results.
Each case should be instantiated with the corresponding logged failure, generated skill/tool
file, visual artifact, and validation record before submission. We use the case studies to make
one specific point: the atomic operations themselves are not new, but \sysname{} can decide from
a failed attempt which operation is needed, generate it as a reusable capability, and validate it
before future use.

\begin{table*}[t]
\centering
\small
\setlength{\tabcolsep}{4pt}
\renewcommand{\arraystretch}{1.15}
\begin{tabular}{p{0.20\textwidth}p{0.24\textwidth}p{0.25\textwidth}p{0.23\textwidth}}
\toprule
\textbf{Case study} & \textbf{Generated capability} & \textbf{Prior motif} & \textbf{Claim supported} \\
\midrule
Case A: Structured-image editing
& A skill that instructs the solver to isolate the relevant chart/table structure, plus a tool
that highlights, boxes, or masks visual evidence.
& ReFocus uses Python visual edits as chain-of-thought for structured image understanding
~\citep{DBLP:conf/icml/FuLYCLY0FZ25}.
& \sysname{} can generate a reusable visual-editing capability from failure, rather than relying
on a fixed visual-editing interface. \\
\midrule
Case B: High-resolution visual search
& A skill that decomposes the question into target-location cues, plus a tool that searches or
zooms into candidate regions.
& ZoomEye performs training-free tree-based image exploration over zoomed sub-regions
~\citep{DBLP:conf/emnlp/ShenZZXZZY25}.
& \sysname{} can generate a local coarse-to-fine search capability without being given ZoomEye as
a predefined module. \\
\midrule
Case C: Interleaved visual skill
& A skill whose SOP includes both text steps and a visual worked example or linked crop from the
triggering failure.
& Visual-thought methods show that intermediate visual artifacts can guide structured reasoning.
& \sysname{} can store procedural and visual evidence together as a retrieved capability. \\
\bottomrule
\end{tabular}
\caption{
  \textbf{Three qualitative forms of evolved capability.}
  The case studies compare generated artifacts to prior visual-reasoning motifs. They do not
  claim that \sysname{} reproduces the full prior algorithms.
}
\label{tab:capability_prior_motifs}
\end{table*}

\subsection{Case A: ReFocus-Style Skill and Tool for Structured Images}
\label{app:refocus_style}

ReFocus~\citep{DBLP:conf/icml/FuLYCLY0FZ25} proposes visual editing as a form of
chain-of-thought for structured image understanding, by asking an MLLM to generate Python
code that edits the input image---drawing boxes, highlighting relevant regions, masking
distractors---before reasoning. The case below illustrates that \sysname{} can produce a
capability with the same flavour \emph{autonomously}: on a ChartQA training case, the
evolution loop diagnosed a visual-disambiguation bottleneck and proposed the paired skill
and tool shown next.

\begin{tcolorbox}[
    title={\textbf{Generated skill (paired with the tool below; triggered on ChartQA)}},
    casebox
]
\begin{lstlisting}[style=mdskill]
## Skill: Edit-Then-Read for Structured Images

When-to-Use:
Use when the question refers to a specific chart/table element that is visually close to distractors, such as one series among many lines, one segment in a stacked bar, or one table row/column.

Strategy:
1. Identify the visual entity named by the question.
2. Before computing the answer, create an edited view that marks only the relevant region and suppresses likely distractors.
3. Re-read the edited image and extract the required value.
4. Perform the requested arithmetic or comparison.

Common Pitfalls:
- Do not answer from the unedited image if multiple marks are visually similar.
- Do not treat the visual edit as adding new information; it only exposes the evidence already present in the image.
\end{lstlisting}
\end{tcolorbox}

\paragraph{Generated tool (paired with the skill above)}
\begin{lstlisting}[style=pycode]
def mark_relevant_region(image_path, region, mode="highlight"):
    """Return an edited image that highlights or masks
    chart/table evidence."""
    from PIL import Image, ImageDraw
    img = Image.open(image_path).convert("RGB")
    draw = ImageDraw.Draw(img, "RGBA")
    x1, y1, x2, y2 = region
    if mode == "highlight":
        draw.rectangle([x1, y1, x2, y2], outline=(255, 0, 0, 255), width=6)
        draw.rectangle([x1, y1, x2, y2], fill=(255, 255, 0, 60))
    elif mode == "mask_distractors":
        mask = Image.new("RGBA", img.size, (255, 255, 255, 180))
        ImageDraw.Draw(mask).rectangle([x1, y1, x2, y2], fill=(0, 0, 0, 0))
        img = Image.alpha_composite( img.convert("RGBA"), mask)
    out_path = image_path.replace(".png", "_marked.png")
    img.convert("RGB").save(out_path)
    return out_path
\end{lstlisting}

Like ReFocus, this capability uses visual editing as an intermediate reasoning step: the
solver does not merely produce a textual explanation but creates an edited visual
artefact that changes the evidence shown to the next solver call. The difference is that
ReFocus defines visual editing as the reasoning interface up front, whereas \sysname{}
emits this skill/tool pair only after the evolution loop's diagnosis identifies visual
disambiguation as the bottleneck on the triggering ChartQA case.

\subsection{Case B: ZoomEye-Style Skill and Tool for High-Resolution Search}
\label{app:zoomeye_style}

ZoomEye~\citep{DBLP:conf/emnlp/ShenZZXZZY25} addresses the loss of fine visual detail by
treating image understanding as a tree-based exploration over zoomed sub-regions: the
full image is the root, zoomed crops are children, and the model searches over promising
regions until it finds the visual evidence needed to answer. The case below illustrates
that \sysname{} can produce a capability with the same flavour \emph{autonomously}: on
an HRBench4K training case where the target object occupied a small fraction of a
high-resolution image, the evolution loop diagnosed a perceptual-resolution bottleneck
and proposed the paired skill and tool shown next.

\begin{tcolorbox}[
    title={\textbf{Generated skill (paired with the tool below; triggered on HRBench4K)}},
    casebox
]
\begin{lstlisting}[style=mdskill]
## Skill: Coarse-to-Fine Region Search

When-to-Use:
Use when the question asks about a small object, distant text, fine-grained attribute, or localized region in a high-resolution image.

Strategy:
1. Phrase the question into a short `target_description` (e.g., "small red sign on the right side of the building").
2. Provide a tile scoring callback `score_tile(tile_path, desc)` that returns a higher value when the tile is more likely to contain the target; in our agent this is a fast VLM yes/no probe.
3. Call `coarse_to_fine_zoom(image_path, target_description, score_tile, grid=2, depth=2)` (defined in the paired tool below). The function recursively splits the image into a 2x2 grid, scores all four tiles, zooms into the best one, and recurses; after `depth=2` it returns the path of the final zoomed crop.
4. Answer the question from the returned zoomed crop. If the crop is still ambiguous, call `coarse_to_fine_zoom` again on the returned crop with `depth=1` to drill in one more level.

Common Pitfalls:
- Do not answer from the globally downsampled image when the requested evidence is small.
- Do not crop solely by image center; rely on the per-tile score and let the function pick.
- Do not use `depth >= 3` without re-checking the crop's resolution; further subdivision yields tiles too small to read text from.
\end{lstlisting}
\end{tcolorbox}

\paragraph{Generated tool (paired with the skill above)}
\begin{lstlisting}[style=pycode]
def coarse_to_fine_zoom(image_path, target_description, score_tile, grid=2, depth=2):
    """Recursively split-and-zoom on an image. At each level the image is divided into a `grid` x `grid` array of tiles, every tile is scored against `target_description` via the supplied `score_tile(tile_path, desc) -> float` callback, and the function recurses into the highest-scoring tile. After `depth` levels the path of the final zoomed crop is returned."""
    from PIL import Image
    crop_path = image_path
    for level in range(depth):
        img = Image.open(crop_path).convert("RGB")
        W, H = img.size
        best_path, best_score = None, -float("inf")
        for gy in range(grid):
            for gx in range(grid):
                box = (gx * W // grid, gy * H // grid, (gx + 1) * W // grid, (gy + 1) * H // grid)
                tile_path = crop_path.replace(".png", f"_L{level}_t{gx}{gy}.png")
                img.crop(box).save(tile_path)
                s = score_tile(tile_path, target_description)
                if s > best_score:
                    best_path, best_score = tile_path, s
        crop_path = best_path
    return crop_path
\end{lstlisting}

Like ZoomEye, this capability avoids forcing the VLM to answer from a single
fixed-resolution full image and instead surfaces candidate sub-regions for a second
solver call. The difference is scope: ZoomEye is a general tree-search algorithm fixed
up front, whereas the \sysname{} capability is produced only after the evolution loop's
diagnosis identifies a perceptual-resolution bottleneck on the triggering HRBench4K case
and implements only the subset of zoom/search behaviour the diagnosis warrants.

\subsection{Case C: Interleaved Visual Skill}
\label{app:interleaved_visual_skills}

The most interesting of the three cases. Here the artefact stored in the library is
\emph{partly visual} and the visual part \emph{teaches by demonstration}: an
\emph{interleaved} skill bundles textual reasoning steps with a pair of reference images,
one \emph{good} example and one \emph{bad} example. The skill body literally instructs
the agent to compare its own intermediate output against the two reference images and
\emph{act differently depending on which one it matches}. The good/bad pair, the
matching text rule, and the diagnostic that produced them are all emitted together by
one evolution iteration and are retrieved together as a bundle on later cases.

\begin{figure}[!t]
\centering
\begin{subfigure}[t]{0.485\columnwidth}
  \centering
  \includegraphics[width=\linewidth]{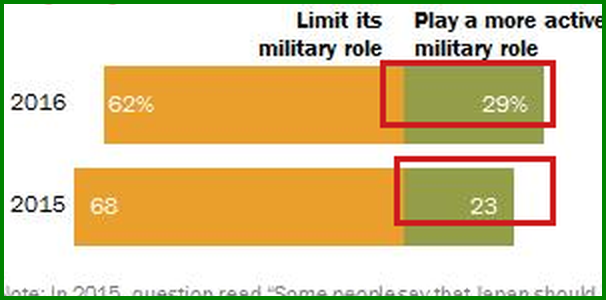}
  \caption{\hlgood{\texttt{crop\_good.png}}: target appearance. Year labels (2016, 2015)
  and category headers are fully visible alongside the bars, so the agent can read off
  values and associate them with the right year.}
  \label{fig:crop_good}
\end{subfigure}\hfill
\begin{subfigure}[t]{0.485\columnwidth}
  \centering
  \includegraphics[width=\linewidth]{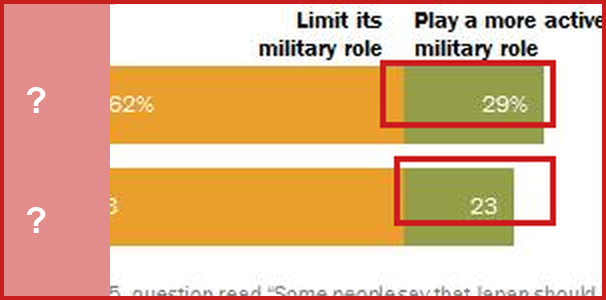}
  \caption{\hlbad{\texttt{crop\_bad.png}}: failure appearance. The year-label column on
  the left is masked out (red overlay with ``?'') to mark what a too-aggressive crop
  would lose. Bars are still visible but the agent can no longer tell which bar is
  which year.}
  \label{fig:crop_bad}
\end{subfigure}
\caption{%
  \textbf{Two reference images stored alongside the Case~C skill.}
  Strategy step~4 of the skill instructs the agent to compare its own intermediate crop
  against these two files: re-do if the crop converges to the right panel, proceed if
  it matches the left panel.
}
\label{fig:chartqa_visual_anchor}
\end{figure}

The triggering case is a ChartQA bar chart where the agent decided to zoom in on the
bars to read their heights more precisely. Its first crop was tight enough that the
\emph{x-axis labels were cut off}: in the zoomed view it could see the bar heights but
could no longer tell which bar corresponded to which year, and answered the wrong year.
The \textsc{AnalyzerDecider} traced the failure to ``\textit{crop too aggressive: target
context lost},'' and the Generator emitted a paired skill+tool whose visual anchor is
\emph{two separate} reference images: \textbf{good-crop}
(Figure~\ref{fig:crop_good}), the same chart re-cropped with $\sim 30$\,px of padding so
the axis labels are fully visible; and \textbf{bad-crop}
(Figure~\ref{fig:crop_bad}), the agent's own failed crop with the axis labels truncated.
The skill text refers to each by name and tells the agent what to do when its own
intermediate crop matches each one.

\begin{tcolorbox}[
    title={\textbf{Generated skill (paired with the tool below; triggered on ChartQA)}},
    casebox
]
\begin{lstlisting}[style=mdskill]
## Skill: Conservative Cropping with Axis-Label Preservation

When-to-Use:
Use when the question requires reading values off a chart and your plan is to crop the chart for higher-resolution inspection. Trigger words: "zoom", "crop", "look closer", "read the y-axis".

Strategy:
1. Identify both the *measurement target* (the bars / lines / points you need to read off) AND the *context* you need to interpret it (x-axis labels, y-axis ticks, legend swatches).
2. Compute a tight bounding box around the measurement target.
3. Expand the bounding box by ~30 px in every direction toward an axis or legend; never crop with margin = 0.
4. Render the crop and compare it against the two reference images stored alongside this skill:
     - If your crop looks like crop_bad.png (axis labels or legend on one side cut off, target visible but no longer associated with its label), the crop is too aggressive. Re-think the bounding box: expand it by another ~30 px in the direction of the missing context and re-crop. Repeat step 4.
     - If your crop looks like crop_good.png (target + axis labels + legend all visible together), proceed to read off values.

Visual References (stored in the skill folder):
- crop_good.png : target appearance for any crop produced under this skill.
- crop_bad.png  : failure appearance; if your crop converges to this shape, redo step 3 with a larger expansion.

Common Pitfalls:
- Cropping with margin = 0 around the bars. This almost always truncates the x-axis labels, which was the failure on the triggering case.
- Assuming the agent can recover from a label-truncated crop by recalling labels from the full image. The full image was the reason for cropping in the first place; the crop is the new source of truth for the agent.
- Cropping too generously (margin >> 50 px), which defeats the purpose of zooming in. Stick to ~30 px.
\end{lstlisting}
\end{tcolorbox}

\paragraph{Generated tool (produces the two reference images used in the skill above)}
\begin{lstlisting}[style=pycode]
def make_crop_reference(image_path, bad_crop_box, expand_px=30):
    """Save the agent's failed crop and a corrected good crop as
    TWO separate reference images for the skill library.
    `bad_crop_box` is the agent's failed crop (axis labels cut off).
    The good crop is the same target with `expand_px` padding
    restored on every side. Returns (good_path, bad_path)."""
    from PIL import Image
    img = Image.open(image_path).convert("RGB")
    W, H = img.size
    l, t, r, b = bad_crop_box
    bad  = img.crop(bad_crop_box)
    good = img.crop((max(0, l - expand_px), max(0, t - expand_px),
                     min(W, r + expand_px), min(H, b + expand_px)))
    bad_path  = image_path.replace(".png", "_crop_bad.png")
    good_path = image_path.replace(".png", "_crop_good.png")
    bad.save(bad_path)
    good.save(good_path)
    return good_path, bad_path
\end{lstlisting}

The interleaved nature here is concrete: the skill is not just text that mentions a
picture, it is a Markdown body whose Strategy step~4 names \emph{two specific stored
images} and prescribes different agent behaviour depending on which one the agent's own
intermediate crop matches. The good reference, the bad reference, the diagnostic
``crop too aggressive'' line, and the conservative-cropping rule were all emitted by
the same evolution iteration on the triggering case, and all four are retrieved together
on later cases of the same problem family. The agent does not have to rediscover the
failure mode or invent the heuristic; it loads them as one bundle.



\end{document}